\documentclass[onecolumn]{article}
\usepackage[latin9]{inputenc}
\usepackage{geometry}
\usepackage{array}
\usepackage{float}
\usepackage{amsbsy}
\usepackage{amstext}
\usepackage{amssymb}
\usepackage{graphicx}
\usepackage[unicode=true,
 bookmarks=true,bookmarksnumbered=true,bookmarksopen=true,bookmarksopenlevel=1,
 breaklinks=false,pdfborder={0 0 0},backref=false,colorlinks=false]
 {hyperref}
\hypersetup{pdftitle={Your Title},
 pdfauthor={Your Name},
 pdfpagelayout=OneColumn, pdfnewwindow=true, pdfstartview=XYZ, plainpages=false}

\makeatletter

\providecommand{\tabularnewline}{\\}
\floatstyle{ruled}
\newfloat{algorithm}{tbp}{loa}
\providecommand{\algorithmname}{Algorithm}
\floatname{algorithm}{\protect\algorithmname}

 \usepackage{algolyx}
 \usepackage{algolyx}

\usepackage[caption=false,font=normalsize,labelfont=sf,textfont=sf]{subfig}
\usepackage{amsbsy}
\usepackage{algorithmic}
\usepackage[english]{babel}

\makeatother

\begin{document}

\title{DeepCare: A Deep Dynamic Memory Model for Predictive Medicine}

\author{Trang~Pham, Truyen~Tran, Dinh~Phung and Svetha~Venkatesh}
\maketitle
\begin{abstract}
Personalized predictive medicine necessitates the modeling of patient
illness and care processes, which inherently have long-term temporal
dependencies. Healthcare observations, recorded in electronic medical
records, are episodic and irregular in time. We introduce DeepCare,
an \emph{end-to-end} deep dynamic neural network that reads medical
records, stores previous illness history, infers current illness states
and predicts future medical outcomes. At the data level, DeepCare
represents care episodes as vectors in space, models patient health
state trajectories through explicit memory of historical records.
Built on Long Short-Term Memory (LSTM), DeepCare introduces time parameterizations
to handle irregular timed events by moderating the forgetting and
consolidation of memory cells. DeepCare also incorporates medical
interventions that change the course of illness and shape future medical
risk. Moving up to the health state level, historical and present
health states are then aggregated through multiscale temporal pooling,
before passing through a neural network that estimates future outcomes.
We demonstrate the efficacy of DeepCare for disease progression modeling,
intervention recommendation, and future risk prediction. On two important
cohorts with heavy social and economic burden -- diabetes and mental
health -- the results show improved modeling and risk prediction accuracy.
\end{abstract}
\global\long\def\ub{\boldsymbol{u}}
\global\long\def\qb{\boldsymbol{q}}
\global\long\def\xb{\boldsymbol{x}}
\global\long\def\bb{\boldsymbol{b}}
\global\long\def\ab{\boldsymbol{a}}
\global\long\def\hb{\boldsymbol{h}}
\global\long\def\Pb{\boldsymbol{P}}
\global\long\def\cb{\boldsymbol{c}}
\global\long\def\fb{\boldsymbol{f}}
\global\long\def\ib{\boldsymbol{i}}
\global\long\def\ob{\boldsymbol{o}}
\global\long\def\gb{\boldsymbol{g}}
\global\long\def\vb{\boldsymbol{v}}
\global\long\def\pb{\boldsymbol{p}}
\global\long\def\zb{\boldsymbol{z}}

\section{Introduction}

When a patient is admitted to hospital, there are two commonly asked
questions: ``what is happening?'' and ``what happens next?'' The
first question refers to the diagnosis of the illness, the second
is about prediction of future medical risk \cite{steyerberg2009clinical}.
While there are a wide array of diagnostic tools to answer the first
question, the technologies are much less advanced in answering the
second \cite{snyderman2003prospective}. Traditionally, this prognostic
question may be answered by experienced clinicians who have seen many
patients, or by clinical prediction models with well-defined and rigorously
collected risk factors. But this is expensive and of limited availability.
Modern electronic medical records (EMRs) promise to offer a fast and
cheap alternative. An EMR typically contains the history of hospital
encounters, diagnoses and interventions, lab tests and clinical narratives.
The wide adoption of EMRs has led to intensified research in building
predictive models from this rich data source in the past few years
\cite{jensen2012mining,tran2014framework,tran_et_al_kais14}.

Answering to prognostic inquiries necessitates modeling patient-level
temporal healthcare processes. An effective modeling must address
four open challenges: (i) \emph{Long-term dependencies} \emph{in healthcare}:
the future illness and care may depend critically on historical illness
and interventions. For example, the onset of diabetes at middle age
remains a risk factor for the rest of the life; cancers may recur
after years; and a previous surgery may prevent certain future interventions.
(ii) \emph{Representation of} \emph{admission}: an admission episode
consists of a variable-size discrete set containing diagnoses and
interventions. (iii) \emph{Episodic recording and irregular timing}:
medical records vary greatly in length, are inherently episodic in
nature and irregular in time \cite{truyen_et_al_kdd13}. The data
is episodic because it is only recorded when the patient visits hospital
and is undergone an episode of care. The episode is often tightly
packed in a short period, typically ranging from a day to two weeks.
The timing of arrivals is largely random. (iv) \emph{Confounding interactions
between disease progression and intervention}: medical records are
a mixture of the course of illness, the developmental and the intervening
processes. In addition to addressing these four challenges, a predictive
system should be \emph{end-to-end} and \emph{generic} so that it can
be deployed on different hospital implementations of EMRs. An end-to-end
system requires minimal or no feature engineering to read medical
records, infer present illness states and predict future outcomes.

Existing methods are poor in handling such complexity. They inadequately
capture variable length \cite{tran2014framework} and ignore the long-term
dependencies \cite{jackson2003multistate,lu2009prospective}. Temporal
models based on Markovian assumption are unable to model temporal
irregularity and have no memory, and thus can completely forget previous
major illness given an irrelevant episode \cite{arandjelovic2015discovering}.
Deep learning, which has recently revolutionized cognitive fields
such as speech recognition, vision and computational linguistics,
holds a great potential in constructing end-to-end systems \cite{lecun2015deep}.
However, its promise to healthcare has not been realized \cite{liang2014deep,tran2015learning,futoma2015comparison}.

To this end, we introduce DeepCare, an \emph{end-to-end} deep dynamic
memory neural network that addresses the four challenges. DeepCare
is built on Long Short-Term Memory (LSTM) \cite{hochreiter1997long,graves2013generating},
a recurrent neural network equipped with \emph{memory cells} to store
experiences. At each time-step, the LSTM reads an input, updates the
memory cell, and returns an output. Memory is maintained through a
\emph{forget gate} that moderates the passing of memory from one time
step to another, and is updated by seeing new input at each time step.
The output is determined by the memory and moderated by an \emph{output
gate}. In DeepCare, the LSTM models the illness trajectory and healthcare
processes of a patient encapsulated in a time-stamped sequence of
admissions. The inputs to the LSTM are information extracted from
admissions. The outputs represent illness states at the time of admission.
\emph{Memory maintenance enables capturing of long-term dependencies},
thus addressing the first challenge. In fact, this capacity has made
LSTM an ideal model for a variety of sequential domains \cite{graves2009novel,graves2013generating,sutskever2014sequence}.
No LSTM has been used in healthcare, however -- one major difficulty
would be the lack of handling of set inputs, irregular timing and
interventions.

Addressing these three drawbacks, DeepCare modifies LSTM in several
ways. For representing admission, which is a set of discrete elements
such as diagnoses and interventions, the solution is to embed these
elements into continuous vector spaces. Vectors of the same type are
then pooled into a single vector. Type-specific pooled vectors are
then concatenated to represent an admission. In that way, \emph{variable-size
admissions are embedded in to continuous distributed vector space}.
The admission vectors then serve as input features for the LSTM. As
the embedding is learnt from data, the model does not rely on manual
feature engineering.

For \emph{irregular timing}, \emph{the forget gate is extended to
be a function of irregular time gap} between consecutive time steps.
We introduce two new forgetting mechanisms: monotonic decay and full
time-parameterization. The decay mimics the natural forgetting when
learning a new concept in human. The parameterization accounts for
more complex dynamics of different diseases over time. The resulting
model is sparse in time and efficient to compute since only observed
records are incorporated, regardless of the irregular time spacing.
Finally, in DeepCare \emph{the confounding interaction between disease
progression and interventions is modeled as follows}. Interventions
influence the output gate of current illness states and the forget
gate that moderates memory carried into the future. As a result, the
illness states (the output) are moderated by past and current interventions.

Once illness states are outputted by the LSTM layer, they are aggregated
through a new time-decayed multiscale pooling strategy. This allows
further handling of time-modulated memory. Finally at the top layer,
pooled illness states are passed through a neural network for estimating
future prognosis. In short, computation steps in DeepCare can be summarized
as

\begin{equation}
P\left(y\mid\ub_{1:n}\right)=P\left(\mbox{nnet}_{y}\left(\mbox{pool}\left\{ \mbox{LSTM}(\ub_{1:n})\right\} \right)\right)\label{eq:model}
\end{equation}
where $\ub_{1:n}$ is the input sequence of admission observations,
$y$ is the outcome of interest (e.g., readmission), $\mbox{nnet}_{y}$
denotes estimate of the neural network with respect to outcome $y$,
and $P$ is probabilistic model of outcomes. Overall, DeepCare is
an end-to-end prediction model that relies on no manual feature engineering,
is capable of reading generic medical records, memorizing a long history,
inferring illness states and predicting the future risk.

We demonstrate our DeepCare on answering a crucial part of the holy
grail question ``what happens next?''. In particular, we demonstrate
our model on \emph{disease progression,} \emph{intervention recommendation}
and \emph{future risk prediction}. Disease progression refers to the
next disease occurrence given the medical history. Intervention recommendation
is about predicting a subset of treatment procedures for the current
diagnoses. Future risk may involve readmission or mortality within
a predefined period after discharge. We note in passing that the forecasting
of future events may be considerably harder than the traditional notion
of classification (e.g., objects/documents categorization) due to
inherent uncertainty in unseen interleaved events. Our experiments
are demonstrated on two datasets of very different nature -- diabetes
(a well-defined chronic condition) and mental health (a diverse mixture
of many acute and chronic conditions). The cohorts were collected
from a large regional hospital in the period of 2002 to 2013. We show
that DeepCare outperforms state-of-the-art baseline classification
methods.

To summarize, through introducing DeepCare, we make four modeling
contributions: (i) handling long-term dependencies in healthcare;
(ii) a novel representation of variable-size admission as fixed-size
continuous vectors; (iii) modeling episodic recording and irregular
timing; and (iv) capturing confounding interactions between disease
progression and intervention. We also contribute to the healthcare
analytics practice by demonstrating the effectiveness of DeepCare
on disease progression, intervention recommendation and medical risk
prediction. Finally, we wish to emphasize that although DeepCare is
designed as predictive model targeted to healthcare, DeepCare can
be applied to other temporal domains with similar data characteristics
(i.e., long-term dependencies, discrete set inputs, irregular timing
and confounding interventions).

The paper is organized as follows. Section \ref{sec:Background} provides
background for Electronic Medical Records, sequential and deep learning
for healthcare. Section \ref{sec:Preliminaries} presents preliminaries
for DeepCare model: Recurrent neural networks, LSTM and learning word
representation. DeepCare is described in Section \ref{sec:DeepCare}
while the experiments and results are reported in Section \ref{sec:Experiments}.
Finally, Section \ref{sec:Discussion-and-Conclusion} discusses further
and concludes the paper.

\section{Background \label{sec:Background}}

\subsection{Electronic medical records (EMRs)\label{sub:EMRs}}

An electronic medical record (EMR) is a digital version of patients
health information. A wide range of information can be stored in EMRs,
such as detailed records of symptoms, data from monitoring devices,
clinicians' observations \cite{paxton2013developing}. EMR systems
store data accurately, decrease the risk of data replication and the
risk of data lost. EMRs are now widely adopted in developed countries
and are increasingly present in the rest of the world. It is expected
that EMRs in hospital help improve treatment quality and reduce healthcare
costs \cite{groves2013big}.

A typical EMR contains information about a sequence of admissions
for a patient. There are two types of admission methods: planned (routine)
and unplanned (emergency). Unplanned admission refers to transfer
from the emergency department. EMRs typically store admitted time,
discharge time, lab tests, diagnoses, procedures, medications and
clinical narratives. Diagnoses, procedures and medications stored
in EMRs are typically coded in standardized formats. Diagnoses are
represented using WHO's ICD (International Classification of Diseases)
coding schemes\footnote{http://apps.who.int/classifications/icd10/browse/2016/en}.
For example, E10 encodes Type 1 diabetes mellitus, E11 encodes Type
2 diabetes mellitus while F32 indicates depressive episode. The procedures
are typically coded in CPT (Current Procedural Terminology) or ICHI
(International Classification of Health Interventions) schemes \footnote{http://www.who.int/classifications/ichi/en/}.
Medication names can be mapped into the ATC (Anatomical Therapeutic
Chemical) scheme \footnote{http://www.whocc.no/atc\_ddd\_index/}.

The wide adoption of EMRs has led to calls for meaningful use \cite{jensen2012mining,weiskopf2013defining}.
One of the most important uses is building predictive models \cite{jensen2012mining,mathias2013development,truyen_et_al_kdd13,tran2014framework,tran_et_al_kais14}.
Like most applications of machine learning, the bottleneck here is
manual feature engineering due to the complexity of the data \cite{hripcsak2013next}\cite{mathias2013development}.
Our DeepCare solves this problem by building an end-to-end system
where features are learnt automatically from data.

\subsection{Sequential models for healthcare}

Although healthcare is inherently episodic in nature, it has been
well-recognized that modeling the entire illness trajectory is important
\cite{granger2006caring}\cite{huang2014similarity}. Nursing illness
trajectory model was popularized by Strauss and Corbin \cite{corbin1991nursing},
but the model is qualitative and imprecise in time \cite{henly2011health}.
Thus its predictive power is very limited.

Electronic medical records (EMRs) offer the quantitative alternative
with precise timing of events. However, EMRs are complex -- they reflect
the interleaving between the illness processes and care processes.
The timing is irregular -- patients only visit hospital when the illness
is beyond a certain threshold, even though the illness may have been
present long before the visit. Existing work that handles such irregularities
includes interval-based extraction \cite{tran2014framework}, but
this method is rather coarse and does not explicitly model the illness
dynamics.

Capturing disease progression has been of great interest \cite{jensen2014temporal,liu2015temporal},
and much effort has been spent on Markov models \cite{jackson2003multistate,wang2014unsupervised}
and dynamic Bayesian networks \cite{orphanou2014temporal}. However,
healthcare is inherently non-Markovian due to the long-term dependencies.
For example, a routine admission with irrelevant medical information
would destroy the effect of severe illness \cite{arandjelovic2015discovering},
especially for chronic conditions. Irregular timing and interventions
have not been adequately modeled. Irregular-time Bayesian networks
\cite{ramati2012irregular} offer a promise, but its power has yet
to be demonstrated. Further, assuming discrete states are inefficient
since the information pathway has only $\mbox{log}(K)$ bits for $K$
states. Our work assumes distributed and continuous states, thus offering
much larger state space.

\subsection{Deep learning for healthcare}

Deep learning is currently at the center of a new revolution in making
sense of a large volume of data. It has achieved great successes in
cognitive domains such as speech, vision and NLP \cite{lecun2015deep}.
To date, deep learning approach to healthcare has been an unrealized
promise, except for several very recent works \cite{liang2014deep,tran2015learning},
where irregular timing is not property modeled. We observe that is
a considerable similarity between NLP and EMR, where diagnoses and
interventions play the role of nouns and modifiers, and an EMR is
akin to a sentence. A major difference is the presence of precise
timing in EMR, as well as the episodic nature. This suggests that
it is possible to extend NLP language models to EMR, provided that
irregular timing and episodicity are properly handled. Our DeepCare
contributes along that line. Going down to the genetic basis of health,
a recent work called DeepFind \cite{alipanahi2015predicting} uses
convolutional networks to detect regular DNA/RNA motifs. This is unlike
DeepCare, where irregular temporal dynamics are modeled.

\section{Preliminaries \label{sec:Preliminaries}}

In this section, we briefly review building blocks for DeepCare, which
will be described fully in Sec.~\ref{sec:DeepCare}.

\subsection{Recurrent neural network \label{sub:Recurrent-Neural-Network}}

\begin{figure}
\begin{centering}
\includegraphics[bb=150bp 150bp 660bp 370bp,clip,width=0.9\columnwidth]{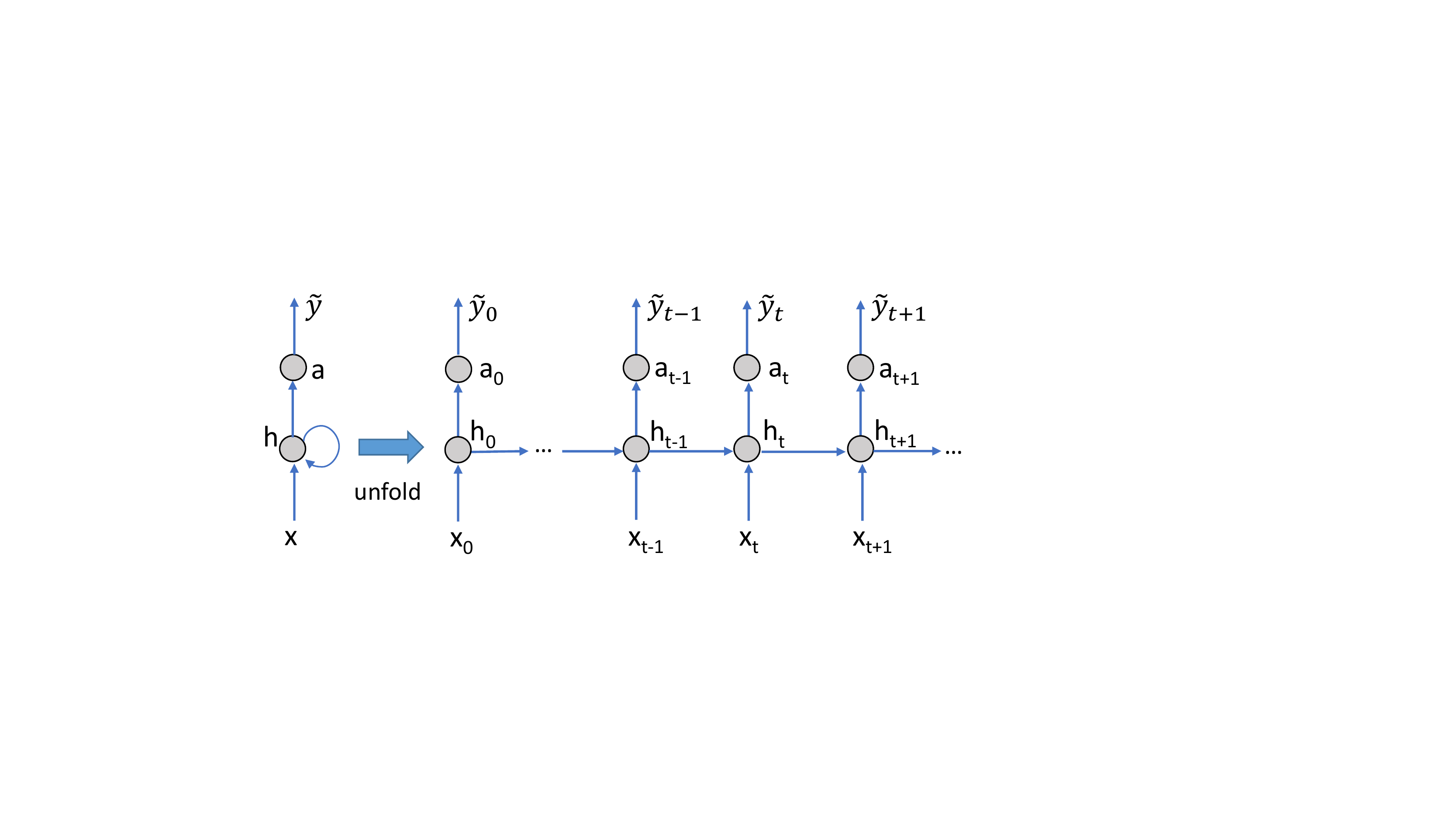} 
\par\end{centering}

\protect\caption{(Left) A typical Recurrent Neural Network and (Right) an RNN unfolded
in time. Each RNN unit at time step $t$ reads input $\protect\xb_{t}$
and previous hidden state $\protect\hb_{t-1}$ , generates output
$\protect\ab_{t}$ and predicts the label $\tilde{y}_{t}$.\label{fig:RNN}}
\end{figure}

A Recurrent Neural Network (RNN) is a neural network repeated over
time. In particular, an RNN allows self-loop connections and shared
parameters across different time steps. While a feedforward neural
network maps an input vector into an output vector, an RNN maps a
sequence into a sequence. Unlike hidden Markov models, where the states
are typically discrete and the transitions between states are stochastic,
RNNs maintain distributed continuous states with deterministic dynamics.
The recurrent connections allow an RNN to memorize previous inputs,
and therefore capture longer dependencies than a hidden Markov model
does. Since the first version of RNN was introduced in the 1980s \cite{rumelhart1986learning},
many varieties of RNN have been proposed such as Time-Delay Neural
Networks \cite{lang1990time} and Echo State Network \cite{jaeger2007echo}.
Here we restrict our discussion to the simple RNN with a single hidden
layer as shown in Fig.~\ref{fig:RNN}.

\subsubsection*{Forward propagation}

An RNN unit has three connections: a recurrent connection from the
previous hidden state to the current hidden state ($\hb_{t-1}\rightarrow\hb_{t}$),
an input-to-hidden-state connection ($\xb_{t}\rightarrow\hb_{t}$)
and a hidden-state-to-output connection ($\hb_{t}\rightarrow\ab_{t}$).
At time step $t$, the model reads the input $\xb_{t}\in\mathbb{R}^{M}$
and previous hidden state $\hb_{t-1}\in\mathbb{R}^{K}$ to compute
the hidden state $\hb_{t}$ (Eq.~\ref{eq:rnn-h}). Thus $\hb_{t}$
summarizes information from all previous inputs $\xb_{0},\xb_{1},...,\xb_{t}$.
The output $\ab_{t}\in\mathbb{R}^{k}$ (Eq.~\ref{eq:rnn-o}) is generated
by a transformation function of $\hb_{t}$, where $k$ is the number
of classes in the classification tasks. To predict the label $\tilde{y}_{t}$,
$\ab_{t}$ is then passed through a probabilistic function $\mbox{f}_{prob}$
to compute the vector of probabilities $\Pb=\left[P\left(\tilde{y}_{t}=0\mid\xb_{t},...,\xb_{0}\right),...,P\left(\tilde{y}_{t}=k-1\mid\xb_{t},...,\xb_{0}\right)\right]$
(Eq.~\ref{eq:rnn-prob}), where $P\left(0\mid\xb_{t},...,\xb_{0}\right),...,P\left(k-1\mid\xb_{t},...,\xb_{0}\right)\geq0$
and $P\left(0\mid\xb_{t},...,\xb_{0}\right)+...+P\left(k-1\mid\xb_{t},...,\xb_{0}\right)=1$.
Denote by $a_{t}^{i}$ the element $i^{th}$ of the vector $\ab_{t}$.
For two classes, $\mbox{f}_{prob}$ is normally a logistic sigmoid
function:

\[
P\left(\tilde{y}_{t}=1\mid\xb_{t},...,\xb_{0}\right)=\mbox{sigmoid}\left(a_{t}^{1}\right)=\frac{1}{1+e^{-a_{t}^{1}}}
\]
and for multiple classes, $\mbox{f}_{prob}$ is a softmax function:

\[
P\left(\tilde{y}_{t}=i\mid\xb_{t},...,\xb_{0}\right)=\mbox{softmax}\left(a_{t}^{i}\right)=\frac{e^{a_{t}^{i}}}{\sum_{j}e^{a_{t}^{j}}}
\]
for $i=0,...,k-1$.

The weighted matrices $W\in\mathbb{R}^{K\times M}$, $U\in\mathbb{R}^{K\times K}$
and $V\in\mathbb{R}^{k\times K}$ and bias vectors $b$ and $c$ are
shared among all time steps. This allows the model to learn with varied
length sequences and produce an output at each time step as follows:

\begin{eqnarray}
\hb_{t} & = & \tanh\left(\bb+W\hb_{t-1}+U\xb_{t}\right)\label{eq:rnn-h}\\
\ab_{t} & = & \cb+V\hb_{t}\label{eq:rnn-o}\\
P\left(\tilde{y}_{t}\right) & = & \mbox{f}_{prob}(\ab_{t})\label{eq:rnn-prob}
\end{eqnarray}
At step $0$, there is no previous hidden state, $\hb_{0}$ is computed
as $\tanh\left(\bb+U\xb_{0}\right)$.

The total loss for a sequence $\xb_{0},\xb_{1},...,\xb_{n}$ and its
corresponding labels $y_{0},y_{1},...,y_{n}$, where $y_{0},y_{1},...,y_{n}\in\left[0,1,..,k-1\right]$,
would be the sum of the losses over all time steps:

\[
L\left(y\mid\xb\right)=\sum_{t=0}^{n}L_{t}\left(\tilde{y}_{t}=y_{t}\mid\xb_{t}...\xb_{0}\right)=-\sum_{t=0}^{n}\log P\left(\tilde{y}_{t}=y_{t}\right)
\]

\subsubsection*{Back-propagation}

RNNs can be trained to minimize the loss function using gradient descent.
The derivatives with respect to the parameters can be determined by
the Back-Propagation Through Time algorithm \cite{werbos1990backpropagation}.
This algorithm obtains the gradients by the chain rule like the standard
back-propagation.

\subsubsection*{Challenge of long-term dependencies}

Many experiments have shown that gradient based learning algorithms
face difficulties in training RNN. This is because the long term dependencies
in long input sequences lead to vanishing or exploding gradients \cite{bengio1994learning,pascanu2012difficulty}.
Many approaches have been proposed to solve the problem, such as Leaky
Units \cite{mozer1993induction}, Nonlinear AutoRegressive models
with eXogenous (NARX) \cite{lin1996learning} and Long-Short Term
Memory (LSTM) \cite{hochreiter1997long}. Among them, LSTM has proved
to be the most effective for handling very long sequences \cite{hochreiter1997long,gers2000learning},
and thus will be chosen as a building block in our DeepCare.

\subsection{Long-short term memory \label{sub:Long-Short-Term-Memory}}

This section reviews Long Short-Term Memory (LSTM) \cite{hochreiter1997long,gers2000learning},
a modified version of RNN, to address the problem of long-term dependencies.
Central to an LSTM is a linear self-loop memory cell which allows
gradients to flow through long sequences. The memory cell is gated
to moderate the amount of information flow into or from the cell.
LSTMs have been significantly successful in many applications, such
as machine translation \cite{sutskever2014sequence}, handwriting
recognition \cite{graves2008unconstrained} and speech recognition
\cite{graves2013speech}.

\begin{figure}
\begin{centering}
\includegraphics[bb=200bp 103bp 620bp 450bp,clip,width=0.6\columnwidth]{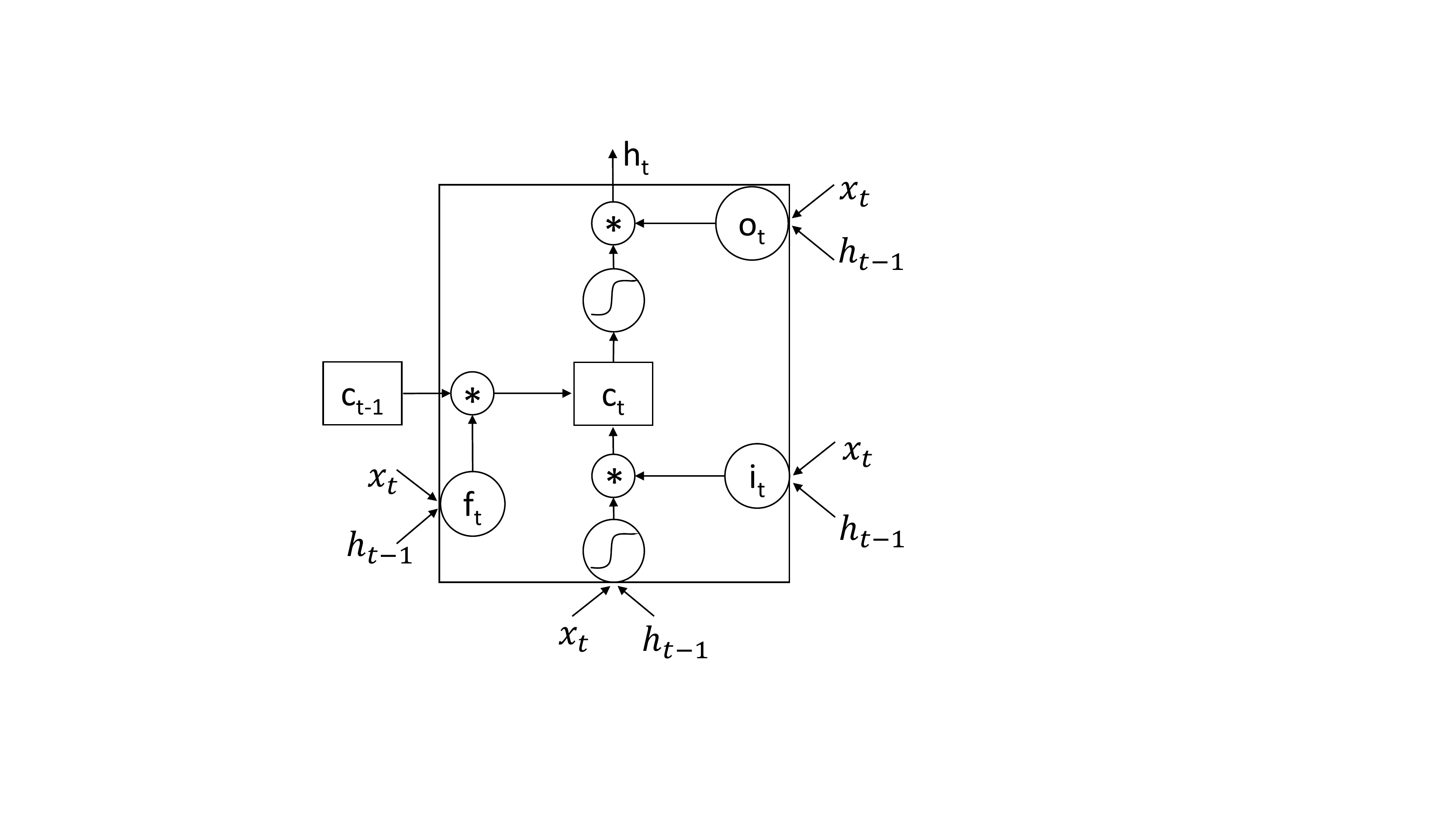} 
\par\end{centering}

\protect\caption{{\small{}{}An LSTM unit that reads input $\protect\xb_{t}$ and previous
output state $\protect\hb_{t-1}$ and produces current output state
$\protect\hb_{t}$. An unit has a memory cell $\protect\cb_{t}$ ,
an input gate $\protect\ib_{t}$, an output gate $\protect\ob_{t}$
and a forget gate $\protect\fb_{t}$.} \label{fig:LSTM-cell}}
\end{figure}

Fig.~\ref{fig:LSTM-cell} describes an LSTM unit. Instead of a simple
RNN unit, an LSTM unit has a memory cell that has state $\cb_{t}\in\mathbb{R}^{K}$
at time $t$. The information flowing through the memory cell is controlled
by three gates: an input gate, a forget gate and an output gate. The
input gate $\ib_{t}\in\mathbb{R}^{K}$ controls the input flowing
into the cell, the forget gate $\fb_{t}\in\mathbb{R}^{K}$ controls
the forgetting of the memory cell, and the output gate $\ob_{t}\in\mathbb{R}^{K}$
moderates the output flowing from the memory cell. Before describing
detailed formulas, we denote the element-wise sigmoid function of
a vector by $\sigma$ and the element-wise product of two vectors
by $*$.

The three gates are all sigmoidal units which set every element of
the gates to a value between 0 and 1:

\begin{eqnarray}
\ib_{t} & = & \sigma\left(W_{i}\xb_{t}+U_{i}\hb_{t-1}+\bb_{i}\right)\label{eq:input-gate}\\
\fb_{t} & = & \sigma\left(W_{f}\xb_{t}+U_{f}\hb_{t-1}+\bb_{f}\right)\label{eq:foget-gate}\\
\ob_{t} & = & \sigma\left(W_{o}\xb_{t}+U_{o}\hb_{t-1}+\bb_{o}\right)\label{eq:output-gate}
\end{eqnarray}
where $W_{\{i,f,o\}}$, $U_{\{i,f,o\}}$, $\bb_{\{i,f,o\}}$ are parameters.
The gates control the amount of information passing through, from
full when the gate value is $1$, to complete blockage when the value
is $0$.

At each time step $t$, the input features are first computed by passing
input $\xb_{t}\in\mathbb{R}^{M}$and the previous hidden state $\hb_{t-1}\in\mathbb{R}^{K}$
through a squashing tanh function:

\begin{equation}
\gb_{t}=\mbox{tanh}\left(W_{c}\xb_{t}+U_{c}\hb_{t-1}+\bb_{c}\right)\label{eq:input-func}
\end{equation}
The memory cell is updated through partially forgetting the previous
memory cell and reading the moderated input features as follows:

\begin{eqnarray}
\cb_{t} & = & \fb_{t}*\cb_{t-1}+\ib_{t}*\gb_{t}\label{eq:mem-update}
\end{eqnarray}
The memory cell sequence is additive, and thus the gradient is also
updated in a linear fashion through the chain rule. This effectively
prevents the gradient from vanishing or exploding. The memory cell
plays a crucial role in memorizing past experiences through the \emph{learnable}
forgetting gates $\fb_{t}$. If $\fb_{t}\rightarrow\mathbf{1}$, all
the past memory is preserved, and new memory keeps updated with new
inputs. If $\fb_{t}\rightarrow\mathbf{0}$, only new experience is
updated and the system becomes memoryless.

Finally, a hidden output state $\hb_{t}$ is computed based on the
memory $\cb_{t}$, gated by the output gate $\ob_{t}$ as follows:

\begin{equation}
\hb_{t}=\ob_{t}*\mbox{tanh}\left(\cb_{t}\right)\label{eq:hidden-LSTM}
\end{equation}
Note that since the system dynamic is deterministic, $\hb_{t}$ is
a function of all previous input: $\hb_{t}=\mbox{LSTM}(\xb_{1:t})$.
The output states are then used to generate outputs. We subsequently
review two output types: sequence labeling and sequence classification.

\subsubsection*{LSTM for sequence labeling}

The output states $\hb_{t}$ can be used to generate labels at time
$t$ as follows: 
\begin{equation}
P\left(y_{t}=l\mid\xb_{1:t}\right)=\mbox{softmax}\left(\vb_{l}^{\top}\hb_{t}\right)\label{eq:LSTM-label-pred}
\end{equation}
for label specific parameters $\vb_{l}$.

\subsubsection*{LSTM for sequence classification}

LSTMs can be used for sequence classification using a simple mean-pooling
strategy over all output states coupled with a differentiable loss
function. For example, in the case of binary outcome $y\in\{0,1\}$,
we have:

\begin{equation}
P\left(y=1\mid\xb_{1:n}\right)=\mbox{LR}\left(\mbox{pool}\left\{ \mbox{LSTM}(\xb_{1:n})\right\} \right)\label{eq:LSTM-seq-class}
\end{equation}
where $\mbox{LR}$ denotes probability estimate of the logistic regression,
and $\mbox{pool}\left\{ \hb_{1:n}\right\} =\frac{1}{n}\sum_{t=1}^{n}\hb_{t}$.

\subsection{Learning word representation \label{sub:Learning-word-representation}}

We use ``word'' to refer to a discrete element within a larger context
(e.g., a word in a document, or a diagnosis in an admission described
in Sec.~\ref{sub:Admission-Embedding}). Recall that input fed into
many machine learning models is often represented as a fix-length
feature vector. For text, bag-of-words are commonly used. A word $w$
is represented by a one-hot vector $\vb_{w}\in\mathbb{R}^{|V|}$,
where $\vb_{w}=\left[v_{w}^{1},...,v_{w}^{|V|}\right]$ and $|V|$
is the number of words in the dictionary: $\vb_{w}=\left[0,...,0,1,0,...,0\right]$
($v_{w}^{i}=1$ if $w=i$, which implies $w$ is the word $i^{th}$
in the dictionary, and $v_{w}^{i}=0$, otherwise). Under bag-of-words
representation, the vector of a sentence $w_{0},...,w_{n}$ is the
sum of its word vectors: $\ub=\vb_{w_{0}}+\vb_{w_{1}}+...+\vb_{w_{n}}$.
However, the bag-of-words method fails to capture ordering and semantic
of the words \cite{le2014distributed}.

A powerful alternative to bag-of-words is to embed words into continuous
distributed representation in a vector space of $M$ dimensions where
$M\ll|V|$ \cite{bengio2003neural}. Every word is map to a unique
vector which is a column in a matrix $E\in\mathbb{R}^{M\times|V|}$.
There are several benefits for word embedding. First, the dimensionality
is greatly reduced and does not depend on the appearance of new words.
Second, the semantic of a word is represented in a distributed fashion,
that is, there are multiple elements that encode the word meaning.
Third, manipulation of continuous vectors is much easier with current
algebraic tools such as addition and matrix multiplication, as evidenced
in recent works \cite{mikolov2013efficient}. For example, the similarity
between two words is simply a cosine between two vectors. More importantly,
the embedding matrix $E$ can be learnt from data.

There are various approaches to learn the embedding matrix $E$. The
most popular approach is perhaps Continuous Bag-of-Words model \cite{mikolov2013efficient}.
For a word $w_{i}$ in a sequence of words, the model uses the words
surrounding $w_{i}$ to predict $w_{i}$. With an input context size
of $C$, $w_{i-C},...,w_{i-1},w_{i+1},...,w_{i+C}$ are called context
words of $w_{i}$. All the context words are embedded into vectors
using embedding matrix $E$ and then averaged to get the mean vector
$\hb$

\[
\hb=\frac{E^{w_{i-C}}+...+E^{w_{i-1}}+E^{w_{i+1}}+..+E^{w_{i+C}}}{2C}
\]
where $E^{t}$ is the column $t^{th}$ of the matrix $E$. The model
then generates the output $\ab=\bar{E}\hb$, where $\bar{E}\in\mathbb{R}^{|V|\times M}$
and predict the center word $w_{i}$ using softmax function

\[
P\left(w_{i}\mid w_{i-C},...,w_{i-1},w_{i+1},...,w_{i+C}\right)=\mbox{softmax}\left(\ab\right)
\]

The parameters $E$ and $\bar{E}$ are learnt by minimizing the loss
function

\[
L=\frac{1}{T}\sum_{i=1}^{T}\log P\left(w_{i}\mid w_{i-C},...,w_{i-1},w_{i+1},...,w_{i+C}\right)
\]
through back-propagation using stochastic gradient descent.

Another approach to learn the embedding matrix $E$ is language modeling
with an RNN \cite{mikolov2010recurrent}. More formally, given a sequence
of words: $w_{0},w_{1},...,w_{t}$, the objective is maximizing the
log probability $\log P\left(w_{t+1}\mid w_{t},...,w_{1},w_{0}\right)$.
Each word $w_{i}$ in the sequence is embedded into vector $\xb_{i}=E^{w_{i}}$
and the sequence $\xb_{0},\xb_{1},...,\xb_{t}$ is the input of an
RNN. The model only produces the output $\ab_{t}$ at the step $t$
(See Sec.~\ref{sub:Recurrent-Neural-Network}, Eq.~\ref{eq:rnn-o})
and predict the next word using a multiclass classifier with a softmax
function

\[
P\left(w_{t+1}\mid w_{t},...,w_{1},w_{0}\right)=\mbox{softmax}\left(\ab_{t}\right)
\]

The loss function is $L=\frac{1}{T}\sum_{t=0}^{T-1}\log p(w_{t+1}\mid w_{t},...,w_{0})$.
The matrix $E$ and all the parameters of the RNN model are learnt
jointly through back-propagation using gradient descent.

\section{DeepCare \label{sec:DeepCare}}

In this section we present our main contribution named DeepCare for
modeling illness trajectories and predicting future outcomes. DeepCare
is built upon LSTM to exploit the ability to model long-term dependencies
in sequences. We extend LSTM to address the three major challenges:
(i) \emph{variable-size discrete inputs}, (ii) \emph{confounding interactions
between disease progression and intervention}, and (iii) \emph{irregular
timing}.

\subsection{Model overview \label{sub:Model-overview}}

Recall from Sec.~\ref{sub:EMRs}, there are two types of admission
methods: planned and unplanned. Let $m_{t}$ be the admission method
at time step $t$, where $m_{t}=1$ indicates unplanned admission
and $m_{t}=2$ indicates planned admission. Let $\Delta t$ be the
elapsed time between the current admission and its previous one.

\begin{figure}[t]
\centering{}\includegraphics[bb=210bp 80bp 690bp 480bp,clip,width=0.8\columnwidth]{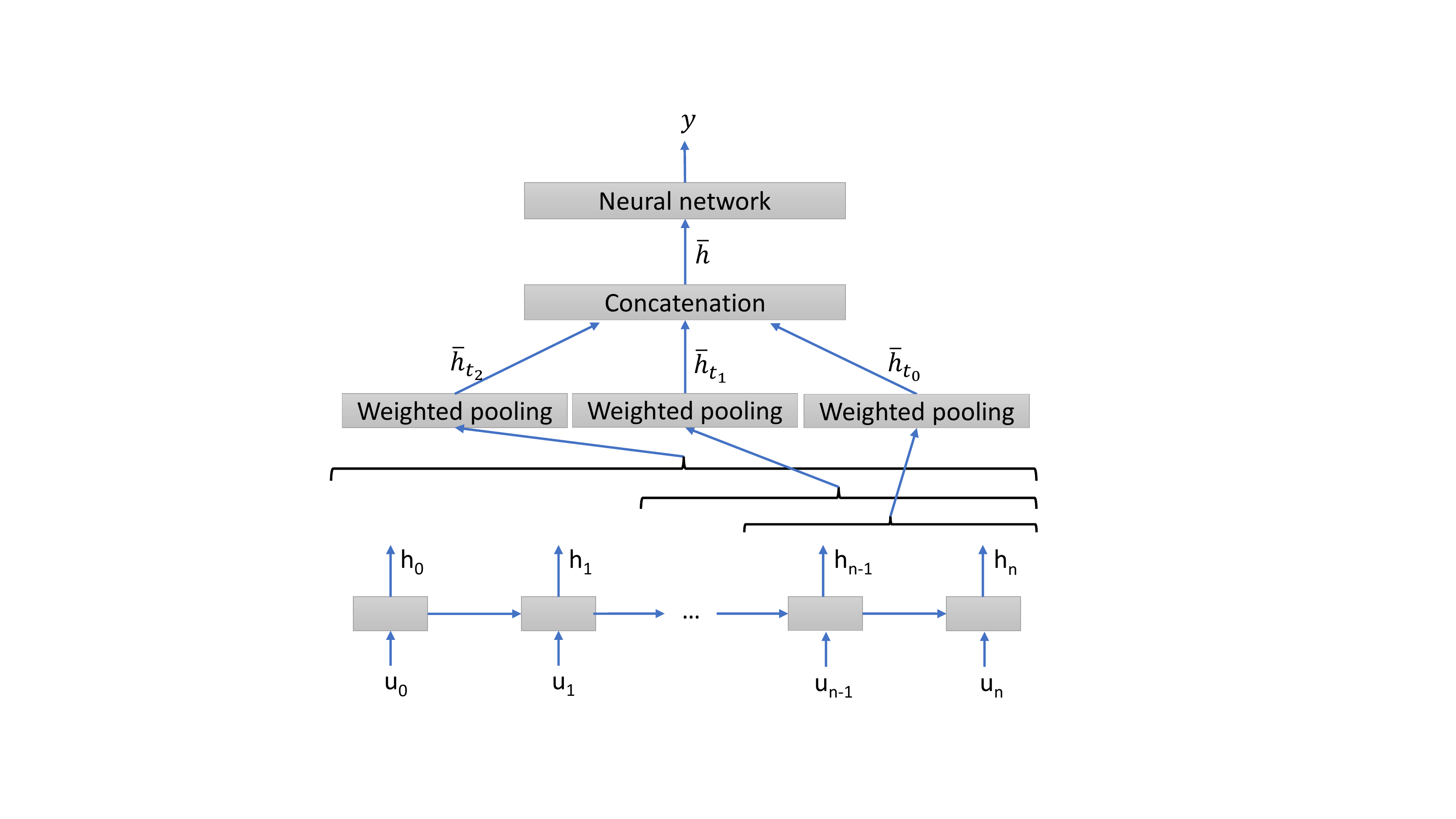}\protect\caption{DeepCare architecture. The bottom layer is Long Short-Term Memory
\cite{hochreiter1997long} with irregular timing and interventions
\label{fig:DeepCare}( see also Fig.~\ref{fig:adm_embed_modified_LSTM}). }
\end{figure}

\begin{figure}
\begin{centering}
\begin{tabular}{cc}
\includegraphics[bb=77bp 120bp 310bp 490bp,clip,width=0.3\columnwidth]{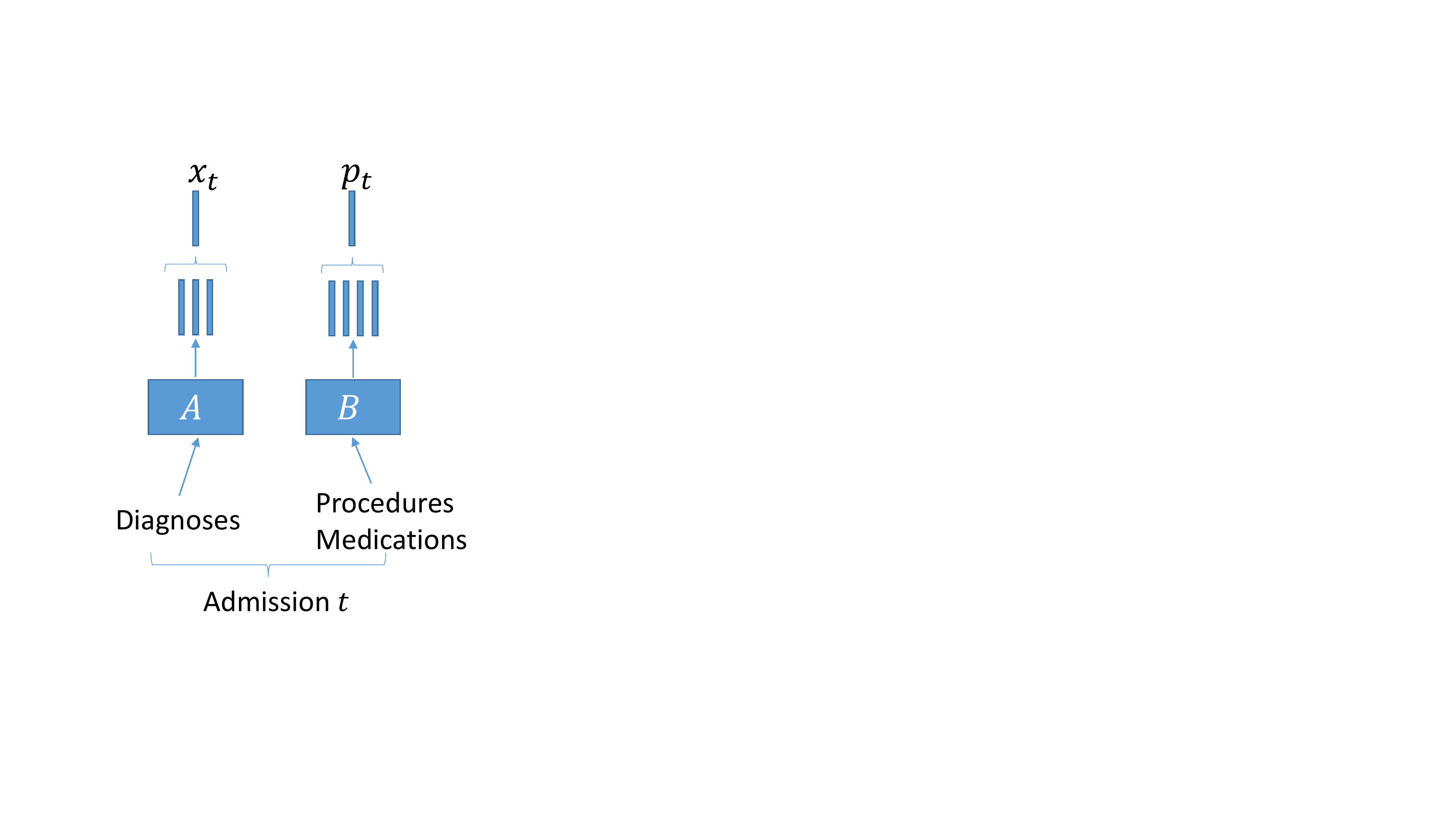}  & \includegraphics[bb=355bp 130bp 780bp 500bp,clip,width=0.4\columnwidth]{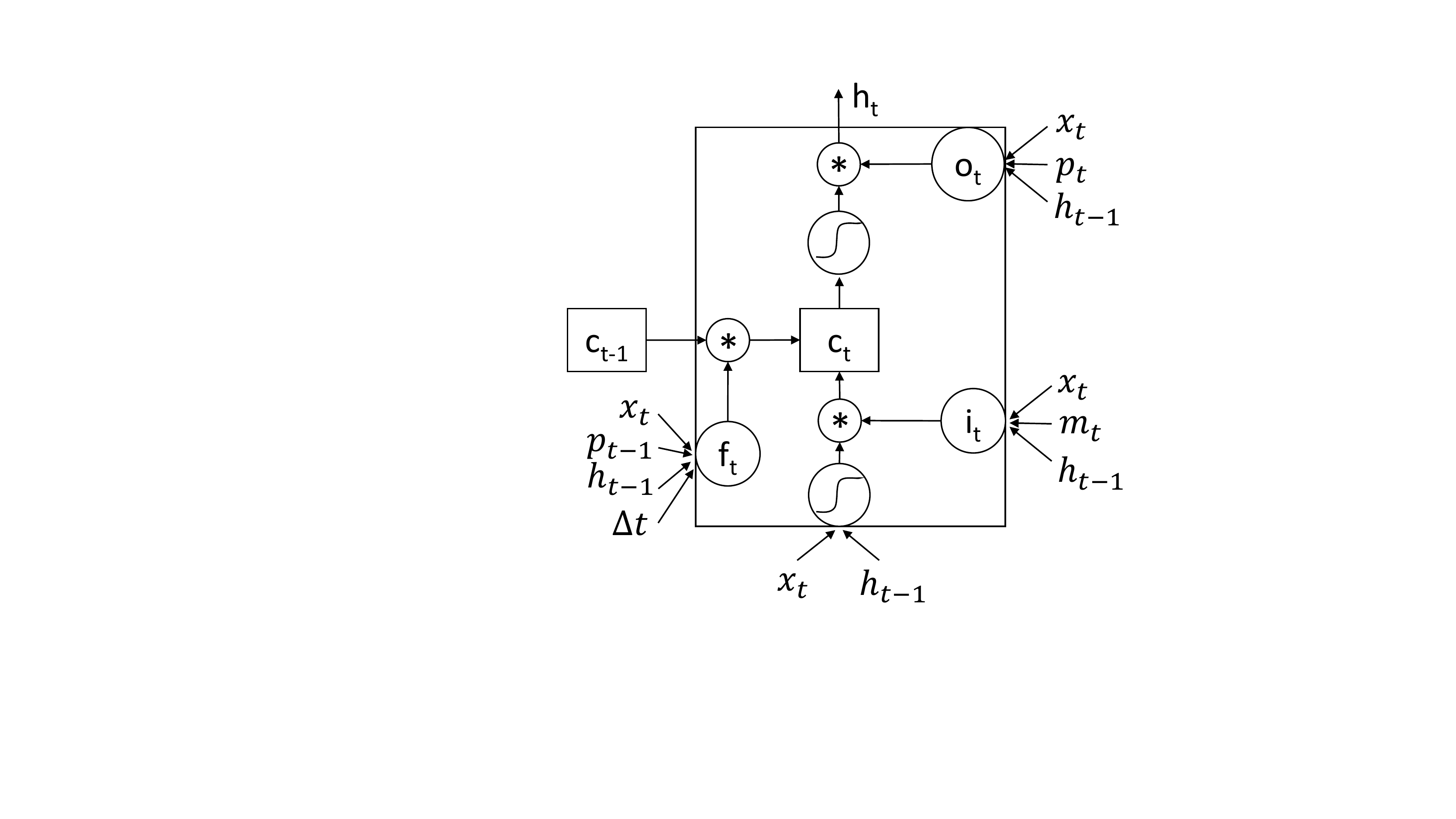}\tabularnewline
(a)  & (b)\tabularnewline
\end{tabular}
\par\end{centering}

\protect\caption{(a) Admission embedding. $A$ and $B$ are embedding matrices. Discrete
diagnoses and interventions are embedded into 2 vectors $\protect\xb_{t}$
and $\protect\pb_{t}$. (b) Modified LSTM unit as a carrier of illness
history. Compared to the original LSTM unit (Fig.~\ref{fig:LSTM-cell}),
the modified unit models times, admission methods, diagnoses and intervention\label{fig:adm_embed_modified_LSTM}}
\end{figure}

As illustrated in Fig.~\ref{fig:DeepCare}, DeepCare is a deep dynamic
neural network that has three main layers. The bottom layer is built
on LSTM whose memory cells are modified to handle irregular timing
and interventions. More specifically, the input is a sequence of admissions.
Each admission $t$ contains a set of diagnosis codes (which is then
formulated as a feature vector $\xb_{t}\in\mathbb{R}^{M}$), a set
of intervention codes (which is further formulated as a feature vector
$\pb_{t}\in\mathbb{R}^{M}$), the admission method $m_{t}\in\text{\{1,2\}}$
and the elapsed time $\Delta t\in\mathbb{R}^{+}$. Denote by $\ub_{0},\ub_{1},...,\ub_{n}$
the input sequence, where $\ub_{t}=\text{[}\xb_{t},\pb_{t},m_{t},\Delta t]$,
the LSTM computes the corresponding sequence of distributed illness
states $\hb_{0},\hb_{1},...,\hb_{n}$, where $\hb_{t}\in\mathbb{R}^{K}$
(See Fig.~\ref{fig:adm_embed_modified_LSTM}b). The middle layer
aggregates illness states through multiscale weighted pooling $\bar{\hb}=\mbox{pool}\left\{ \hb_{0},\hb_{1},...,\hb_{n}\right\} $,
where $\bar{\hb}\in\mathbb{R}^{sK}$ for $s$ scales.

The top layer is a neural network that takes pooled states and other
statistics to estimate the final outcome probability, as summarized
in Eq.~(\ref{eq:model}) as

\[
P\left(y\mid\ub_{0:n}\right)=P\left(\mbox{nnet}_{y}\left(\mbox{pool}\left\{ \mbox{LSTM}(\ub_{0:n})\right\} \right)\right)
\]
The probability $P\left(y\mid\ub_{0:n}\right)$ depends on the nature
of outputs and the choice of statistical structure. For example, for
binary outcome, $P\left(y=1\mid\ub_{0:n}\right)$ is a logistic function;
for multiclass outcome, $P\left(y\mid\ub_{0:n}\right)$ is a softmax
function; and for continuous outcome, $P\left(y\mid\ub_{0:n}\right)$
is a Gaussian. In what follows, we describe the first two layers in
more detail.

\subsection{Representing variable-size admissions \label{sub:Admission-Embedding}}

There are two main types of information recorded in an admission:
(i) diagnoses of current condition; and (ii) interventions. Interventions
include procedures and medications. Diagnoses, procedures and medications
are coded using coding schemes which are described in Sec.~\ref{sub:EMRs}.
These schemes are hierarchical and the vocabularies are of tens of
thousands in size. Thus for a problem, a suitable coding level should
be used for balancing between specificity and robustness.

Our approach is to embed admissions into vectors. Fig.~\ref{fig:adm_embed_modified_LSTM}a
illustrates the embedding method. An admission is a set of a varied
number of codes (diagnoses and interventions). Codes are first embedded
into vectors, analogous to word embedding described in Sec.~\ref{sub:Learning-word-representation}.
We then pool all the present diagnosis vectors to derive $\xb_{t}\in\mathbb{R}^{M}$.
Likewise, we derive the pooled intervention vector $\pb_{t}\in\mathbb{R}^{M}$.
Finally, an admission embedding is a $2M$-dim vector $\left[\xb_{t},\pb_{t}\right]$.

\subsubsection{Pooling \label{sub:Pooling}}

Let $\mathcal{D}$ be the set of diagnosis codes and $\mathcal{I}$
be the set of intervention codes. The two sets are indexed from 1
to $|\mathcal{D}|$ and from 1 to $|\mathcal{I}|$, respectively.
Denote diagnosis embedding matrix by $A\in\mathbb{R}^{M\times|\mathcal{D}|}$
and intervention embedding matrix by $B\in\mathbb{R}^{M\times|\mathcal{I}|}$.
Let $A^{j}$ is the $j^{th}$ column and $A_{i}^{j}$ is the element
at the $j^{th}$column and the $i^{th}$ row of the matrix $A$. Let
$x_{t}^{i}$ be the $i^{th}$ element of the vector $\xb_{t}$ and
$p_{t}^{i}$ be the $i^{th}$ element of the vector $\pb_{t}$. Each
admission $t$ contains $h$ diagnoses: $d_{1},d_{2},...,d_{h}\in\{1,2,..,|\mathcal{D}|\}$
and $k$ interventions: $s_{1},s_{2},...,s_{k}\in\{1,2,...,|\mathcal{I}|\}$.
The admission is pooled by max, sum or mean pooling as follow:
\begin{itemize}
\item \emph{Max pooling admission (max adm.)}. The pooling is element-wise
as follows:
\[
\xb_{t}^{i}=\max\left(A_{i}^{d_{1}},A_{i}^{d_{2}},...,A_{i}^{d_{h}}\right)
\]
\[
\pb_{t}^{i}=\max\left(B_{i}^{s_{1}},B_{i}^{s_{2}},...,B_{i}^{s_{k}}\right)
\]
for $i=1,...,M$. This is analogous to paying selective attention
to the element of the highest impact among diagnoses and among interventions.
It also resembles the usual coding practice that one diagnosis is
picked as the primary reason for admission.
\item \emph{Normalized sum pooling admission (sum adm.)}. In healthcare,
risk loosely adds up. A patient with multiple diseases (multiple comorbidities)
is more likely to be at risk than those with single condition. We
propose the following normalized sum pooling method:
\[
\xb_{t}^{i}=\frac{A_{i}^{d_{1}}+A_{i}^{d_{2}}+...+A_{i}^{d_{h}}}{\sqrt{\mid A_{i}^{d_{1}}+A_{i}^{d_{2}}+...+A_{i}^{d_{h}}\mid}}
\]
\[
\pb_{t}^{i}=\frac{B_{i}^{s_{1}}+B_{i}^{s_{2}}+...+B_{i}^{s_{k}}}{\sqrt{\mid B_{i}^{s_{1}}+B_{i}^{s_{2}}+...+B_{i}^{s_{k}}\mid}}
\]
for $i=1,...,M$. The normalization is to reduce the effect of highly
variable length.
\item \emph{Mean pooling admission (mean adm.)}. In absence of primary conditions,
a mean pooling could be a sensible choice:
\[
\xb_{t}=\frac{A^{d_{1}}+A^{d_{2}}+...+A^{d_{h}}}{h}
\]
\[
\pb_{t}=\frac{B^{s_{1}}+B^{s_{2}}+...+B^{s_{k}}}{k}
\]

\end{itemize}

\subsubsection{Admission as input}

Once admission embedding has been derived, diagnosis embedding is
used as input for the LSTM. As interventions are designed to reduce
illness, their effect is modeled separately in Sec.~\ref{sub:Modeling-effect-of-intervention}.
There are two main types of admission: planned and unplanned. Unplanned
admissions refer to transfer from emergency attendances, which typically
indicates higher risk. Recall from Eqs.~(\ref{eq:input-gate},\ref{eq:input-func})
that the input gate $\ib$ control how much new information is updated
into memory $\cb$. The gate can be modified to reflect the risk level
of admission type as follows:

\begin{equation}
\ib_{t}=\frac{1}{m_{t}}\sigma\left(W_{i}\xb_{t}+U_{i}\hb_{t-1}+\bb_{i}\right)\label{eq:modified-input-gate}
\end{equation}
where $m_{t}=1$ if the admission method is unplanned, $m_{t}=2$
otherwise, and $\sigma$ is the element-wise sigmoid function of a
vector.

\subsection{Modeling effect of interventions \label{sub:Modeling-effect-of-intervention}}

The intervention vector ($\pb_{t}$) of an admission is modeled as
illustrated in Fig.~\ref{fig:adm_embed_modified_LSTM}b. Since interventions
are designed to cure diseases or reduce patient's illness, the output
gate, which controls the illness states, is moderated by the \emph{current}
intervention as follows: 
\begin{equation}
\ob_{t}=\sigma\left(W_{o}\xb_{t}+U_{o}\hb_{t-1}+P_{o}\pb_{t}+\bb_{o}\right)\label{eq:modified-output-gate}
\end{equation}
where $P_{o}$ is the intervention weight matrix for the output gate
and $\pb_{t}$ is intervention at time step $t$.

Moreover, interventions may have long-term impacts (e.g., curing disease
or introducing toxicity). This suggests the illness forgetting is
moderated by \emph{previous} intervention 
\begin{equation}
\fb_{t}=\sigma\left(W_{f}\xb_{t}+U_{f}\hb_{t-1}+P_{f}\pb_{t-1}+\bb_{f}\right)\label{eq:forget-time-intervention}
\end{equation}
where $\pb_{t-1}$is intervention embedded vector at time step $t-1$
and $P_{f}$ is the intervention weight matrix for the forget gate.

\subsection{Capturing time irregularity \label{sub:Capturing-time-irregularity}}

When a patient's history is modeled by LSTM (Sec.~\ref{sub:Long-Short-Term-Memory}),
the memory cell carries the illness history. But this memory needs
not be constant as illness states change over time. We introduce two
mechanisms of forgetting the memory by modified the forget gate $\fb_{t}$
in Eq.~\ref{eq:forget-time-intervention}:

\subsubsection{Time decay \label{sub:Time-decay}}

There are acute conditions that naturally reduce their effect through
time. This suggests a simple decay modeled in the forget gate $\fb_{t}$:

\begin{equation}
\fb_{t}\leftarrow d\left(\Delta_{t-1:t}\right)\fb_{t}\label{eq:time-decay}
\end{equation}
where $\Delta_{t-1:t}$ is the time passed between step $t-1$ and
step $t$, and $d\left(\Delta_{t-1:t}\right)\in(0,1]$ is a decay
function, i.e., it is monotonically non-increasing in time. One function
we found working well is $d(\Delta_{t-1:t})=\left[\log(e+\Delta_{t-1:t})\right]^{-1}$,
where $\Delta_{t-1:t}$ is measured in days and $e\approx2.718$ is
the the base of the natural logarithm.

\subsubsection{Parametric time \label{sub:Parametric-time}}

Time decay may not capture all conditions, since some conditions can
get worse, and others can be chronic. This suggests a more flexible
parametric forgetting: 
\begin{equation}
\fb_{t}=\sigma\left(W_{f}\xb_{t}+U_{f}\hb_{t-1}+Q_{f}\qb_{\Delta_{t-1:t}}+P_{f}\pb_{t-1}+\bb_{f}\right)\label{eq:time-parameterize}
\end{equation}
where $\qb_{\Delta_{t-1:t}}$ is a vector derived from the time difference
$\Delta_{t-1:t}$, $Q_{f}$ is the parametric time weight matrix.
For example, we may have: $\qb_{\Delta_{t-1:t}}=\left(\frac{\Delta_{t-1:t}}{60},\left(\frac{\Delta_{t-1:t}}{180}\right)^{2},\left(\frac{\Delta_{t-1:t}}{365}\right)^{3}\right)$
to model the third-degree forgetting dynamics. $\Delta_{t-1:t}$ is
measured in days and is divided by 60, 180 and 365 to prevent the
vector $\qb_{\Delta t-1:t}$ from large values.

\subsection{Prognosis through multiscale pooling and recency attention \label{sub:Multiscale-pooling}}

Once the illness dynamics have been modeled using the memory LSTM,
the next step is to aggregate the illness states to infer about the
future prognosis (Fig.~\ref{fig:DeepCare}). The simplest way is
to use mean-pooling, where $\bar{\hb}=\mbox{pool}\left\{ \hb_{0:n}\right\} =\frac{1}{n+1}\sum_{t=0}^{n}\hb_{t}$.
However, this does not reflect the attention to recency in healthcare.
Here we introduce a simple attention scheme that weighs recent events
more than old ones: $\bar{\hb}=\left(\sum_{t=t_{0}}^{n}r_{t}\hb_{t}\right)/\sum_{t=t_{0}}^{n}r_{t},$
where

\begin{eqnarray*}
r_{t} & = & \left[m_{t}+\mbox{log}\left(1+\Delta_{t:n}\right)\right]^{-1}
\end{eqnarray*}
and $\Delta_{t:n}$ is the elapsed time between the step $t$ and
the current step $n$, measured in months; $m_{t}=1$ if emergency
admission, $m_{t}=2$ if routine admission. The starting time step
$t_{0}$ is used to control the length of look-back in the pooling,
for example, $\Delta_{t_{0}:n}\le12$ for one year look-back. Since
diseases progress at different rates for different patients, we employ
multiple look-backs: $12$ months, $24$ months, and all available
history. Finally, the three pooled illness states are stacked into
a vector: $\bar{\hb}=\left[\bar{\hb}_{12},\bar{\hb}_{24},\bar{\hb}_{all}\right]$
which is then fed to a neural network for inferring about future prognosis.

\subsection{Model complexity}

The number of model parameters are $M\times|V|+M\times K+K\times K+K\times D$,
which consists of the following components:

\subsubsection*{\emph{Parameters in the LSTM layer}}
\begin{itemize}
\item For admission embedding, we use two embedding matrices $A$ and $B$.
We have $A+B\in\mathbb{R}^{M\times|V|}$ 
\item The input gate: $W_{i}\in\mathbb{R}^{M\times K}$, $U_{i}\in\mathbb{R}^{K\times K}$
and $b_{i}\in\mathbb{\mathbb{R}}^{K\times1}$ 
\item The output gate: $W_{o}\in\mathbb{R}^{M\times K}$, $U_{o}\in\mathbb{R}^{K\times K}$,
$P_{o}\in\mathbb{R}^{K\times K}$ and $b_{o}\in\mathbb{\mathbb{R}}^{K\times1}$ 
\item The forget gate: $W_{f}\in\mathbb{R}^{M\times K}$, $U_{f}\in\mathbb{R}^{K\times K}$,
$P_{f}\in\mathbb{R}^{K\times K}$ and $b_{f}\in\mathbb{\mathbb{R}}^{K\times1}$.
In the case of time decay there are no other parameters and in the
case of parametric time, the forget gate has a time weight matrix
$Q_{f}\in\mathbb{R}^{N_{time}\times K}$ ($N_{time}=3$ in our implementation) 
\item The memory cell: $W_{i}\in\mathbb{R}^{M\times K}$, $U_{i}\in\mathbb{R}^{K\times K}$
and $b_{i}\in\mathbb{\mathbb{R}}^{K\times1}$ 
\end{itemize}

\subsubsection*{\emph{Parameters in the Neural network layer}}
\begin{itemize}
\item The neural network layer consists of an input-hidden weight matrix
$U_{h1}\in\mathbb{R}^{3K\times D}$, hidden-output weight matrix $U_{h2}\in\mathbb{R}^{D\times2}$
and two bias vectors $c_{1}\in\mathbb{R}^{D\times1}$ and $c_{2}\in\mathbb{R}^{2x1}$ 
\end{itemize}

\subsection{Learning \label{sub:Learning}}

Once all the illness states are pooled and stacked into vector $\bar{\hb}$,
$\bar{\hb}$ is then fed to a neural network with one hidden layer

\begin{eqnarray}
\ab_{h} & = & \sigma\left(U_{h}\bar{\hb}+\bb_{h}\right)\label{eq:activation-hidden}\\
\zb_{y} & = & U_{y}\ab_{h}+\bb_{y}\label{eq:z_y}\\
P\left(y\mid\ub_{1:n}\right) & = & \mbox{f}_{prob}\left(\zb_{y}\right)\label{eq:prob-nnet}
\end{eqnarray}

Learning is carried out through minimizing cross-entropy: $L=-\log P\left(y\mid\ub_{0:n}\right)$.
For example, in the case of binary classification, $y\in\{0,1\}$,
we use logistic regression to represent $P\left(y\mid\ub_{0:n}\right)$,
i.e. $P\left(y=1\mid\ub_{0:n}\right)=\sigma\left(\zb_{y}\right)$.
The cross-entropy becomes

\begin{equation}
L=-y\log\sigma-\left(1-y\right)\log\left(1-\sigma\right)\label{eq:Loss}
\end{equation}

Despite having a complex structure, DeepCare's loss function is fully
differentiable, and thus can be minimized using standard back-propagation.
The learning complexity is linear with the number of parameters. See
Alg.~\ref{alg:DeepCare-forward-pass} for an overview of DeepCare
forward pass.

\begin{algorithm}

\caption{DeepCare forward pass \label{alg:DeepCare-forward-pass}}

\begin{algor}
\item [{Input:}] \textbf{Inputs: }Patients' disease history records\end{algor}
\begin{algor}[1]
\item [{for}] each step $t$
\item [{{*}}] $[\xb_{t},\pb_{t}]=\mbox{embedding}(d_{1},...,d_{h},s_{1},...,s_{k})$
(Sec.~\ref{sub:Pooling})
\item [{{*}}] Compute 3 gates: $\ib_{t}$ (Eq.~\ref{eq:modified-input-gate}),
$\ob_{t}$ (Eq.~\ref{eq:modified-output-gate}), $\fb_{t}$ (Eq.~\ref{eq:time-decay}
or Eq.~\ref{eq:time-parameterize})
\item [{{*}}] Compute $\cb_{t}$ (Eq.~\ref{eq:mem-update}) and $\hb_{t}$
(Eq.~\ref{eq:hidden-LSTM})
\item [{endfor}]~
\item [{{*}}] Compute $\bar{\hb}$ (Sec.~\ref{sub:Multiscale-pooling})
\item [{{*}}] Compute $P(y\mid\ub_{0:n})$ (Eq.~\ref{eq:activation-hidden},\ref{eq:z_y},\ref{eq:prob-nnet})
\item [{{*}}] Compute loss function $L$ (Eq.~\ref{eq:Loss})
\end{algor}
\end{algorithm}

\subsection{Pretraining and regularization}

\subsubsection{Pretraining with auxiliary tasks}

Pretraining can be done by unsupervised learning on unlabeled data
\cite{hinton2006fast,dai2015semi}. Pretraining has been proven to
be effective because it helps the optimization by initializing weights
in a region near a good local minimum \cite{bengio2007greedy,erhan2010does}.
In our work we use auxiliary tasks to pretrain the model for future
risk prediction tasks. In our case, auxiliary tasks are predicting
diagnoses of the next readmission and predicting interventions of
current admission. These tasks play a role in disease progression
tracking and intervention recommendation.

We use the bottom layer of DeepCare for training auxiliary tasks.
As described in Sec.~\ref{sub:Model-overview}, the LSTM layer reads
a sequence of admissions $\ub_{0},\ub_{1},...,\ub_{n}$ and computes
the corresponding sequence of distributed illness states $\hb_{0},\hb_{1},...,\hb_{n}$.
At each step $t$, $\hb_{t}$ is used to generate labels $y_{t}$
by the formula given in Eq.~(\ref{eq:LSTM-label-pred}) where $y_{t}$
can be a set of diagnoses or interventions. After training, the code
embedding matrix is then used to initialize the embedding matrix for
training the risk prediction tasks. The results of next readmission
diagnosis prediction and current admission intervention prediction
are reported in Sec.~\ref{sub:Disease progression} and Sec.~\ref{sub:Intervention-recommendation}.

\subsubsection{Regularization}

DeepCare may lead to overfitting because it introduces three more
parameter matrices to the sigmoid gates to handle interventions and
time. Therefore, we use L2-norm and Dropout to prevent overfitting.
L2-norm regularization, also called ``weight decay'', is used to
prevent weight parameters from extreme values. A constant $\lambda$
is introduced to control the magnitude of the regularization. Dropout
is a regularization method for DNNs. During training, units are deleted
with a pre-defined probability $1-p$ (dropout ratio) and the remaining
parts are trained through back-propagation as usual \cite{srivastava2014dropout,baldi2013understanding}.
This prevents the co-adaptation between units, and therefore prevents
overfitting. At the test time, a single neural net is used without
dropout and the outgoing weights of a unit that is retained with probability
$p$ during training are multiplied by $p$. This combines $2^{k}$($k$
is the number of units) shared weight networks into a single neural
network at test time. Therefore, dropout is also considered as an
ensemble method.

However, the original version of dropout does not work well with RNNs
because it may hurt the dependencies in sequential data during training
\cite{bayer2013fast,zaremba2014recurrent}. Thus, dropout in DeepCare
is only introduced at input layer and neural network layer: 
\begin{itemize}
\item \emph{Dropout codes:} Before pooling the embedding vectors of diagnoses
and interventions in each admission, each of these embedding vectors
is deleted with probability $1-p_{code}$ 
\item \emph{Dropout input features:} After deriving $[\xb_{t},\pb_{t}]$
as described in Sec.~\ref{sub:Admission-Embedding}, each value in
these two vector is dropped with probability $1-p_{feat}$ 
\item \emph{Dropout units in neural network layer:} The pooled state $z$
as described in Sec.~\ref{sub:Admission-Embedding} is feed as the
input of the neural network. Dropout is used at input units with probability
$1-p_{in}$ and at hidden units with probability $1-p_{hidd}$.\end{itemize}

\section{Experiments \label{sec:Experiments}}

We model disease progression, intervention recommendation and future
risk prediction in two very diverse cohorts: mental health and diabetes.
These diseases differ in causes and progression.

\begin{figure*}
\begin{centering}
\begin{tabular}{>{\raggedright}p{0.3\textwidth}>{\raggedright}p{0.3\textwidth}>{\raggedright}p{0.3\textwidth}}
(a) Age  & (b) Admission  & (c) Length of stay (days)\tabularnewline
\includegraphics[width=0.3\textwidth]{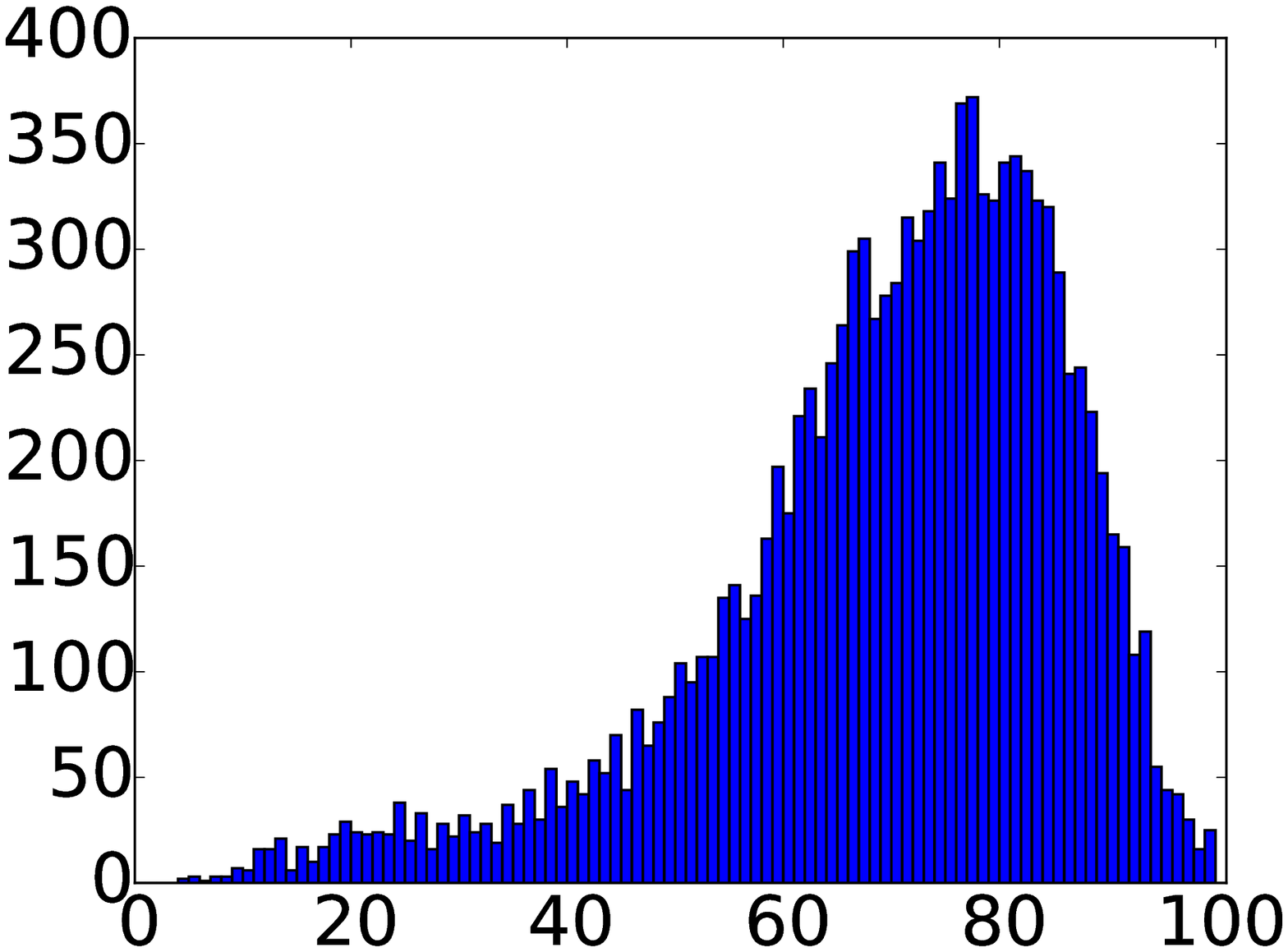}  & \includegraphics[width=0.3\textwidth]{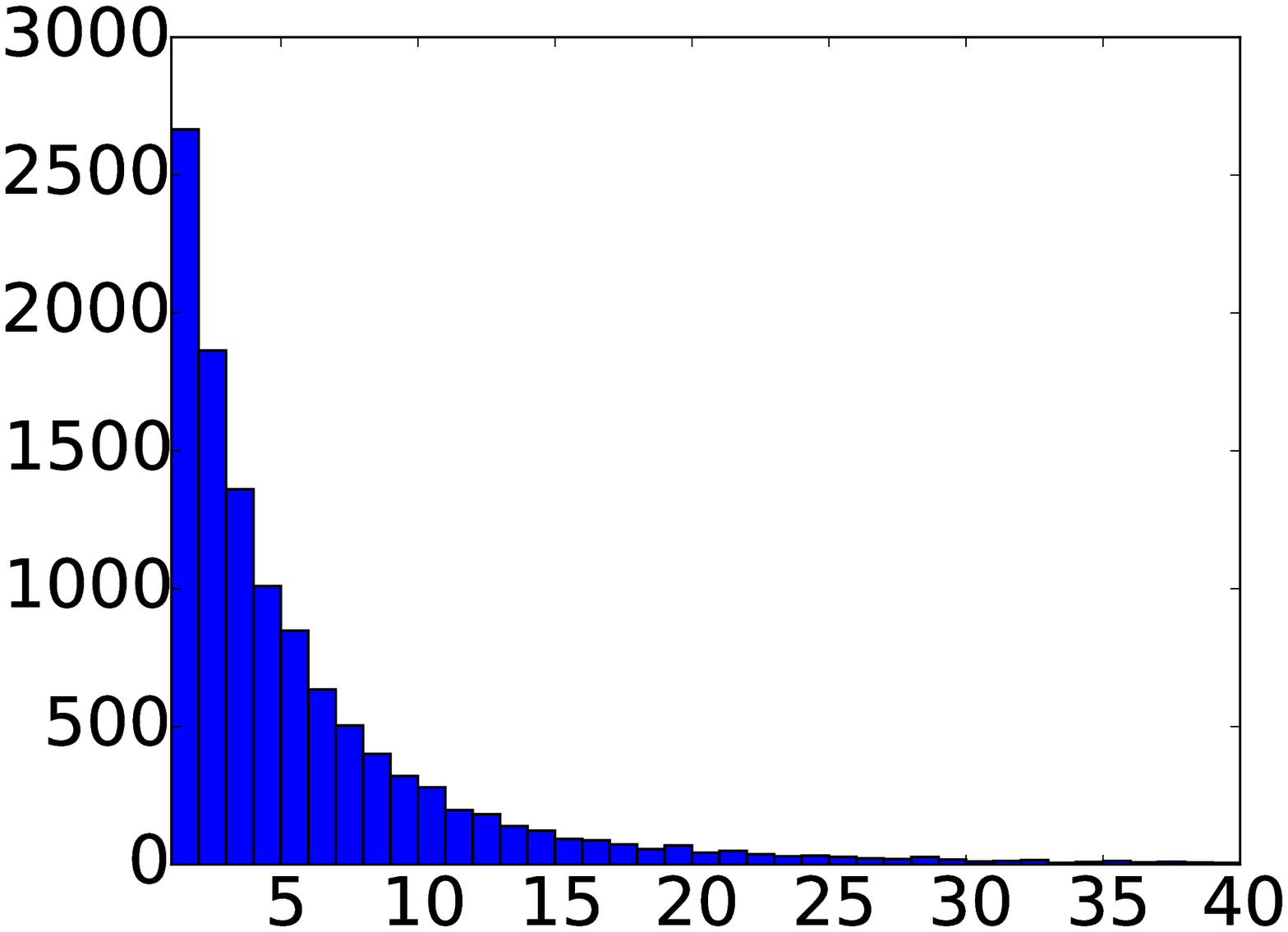}  & \includegraphics[width=0.3\textwidth]{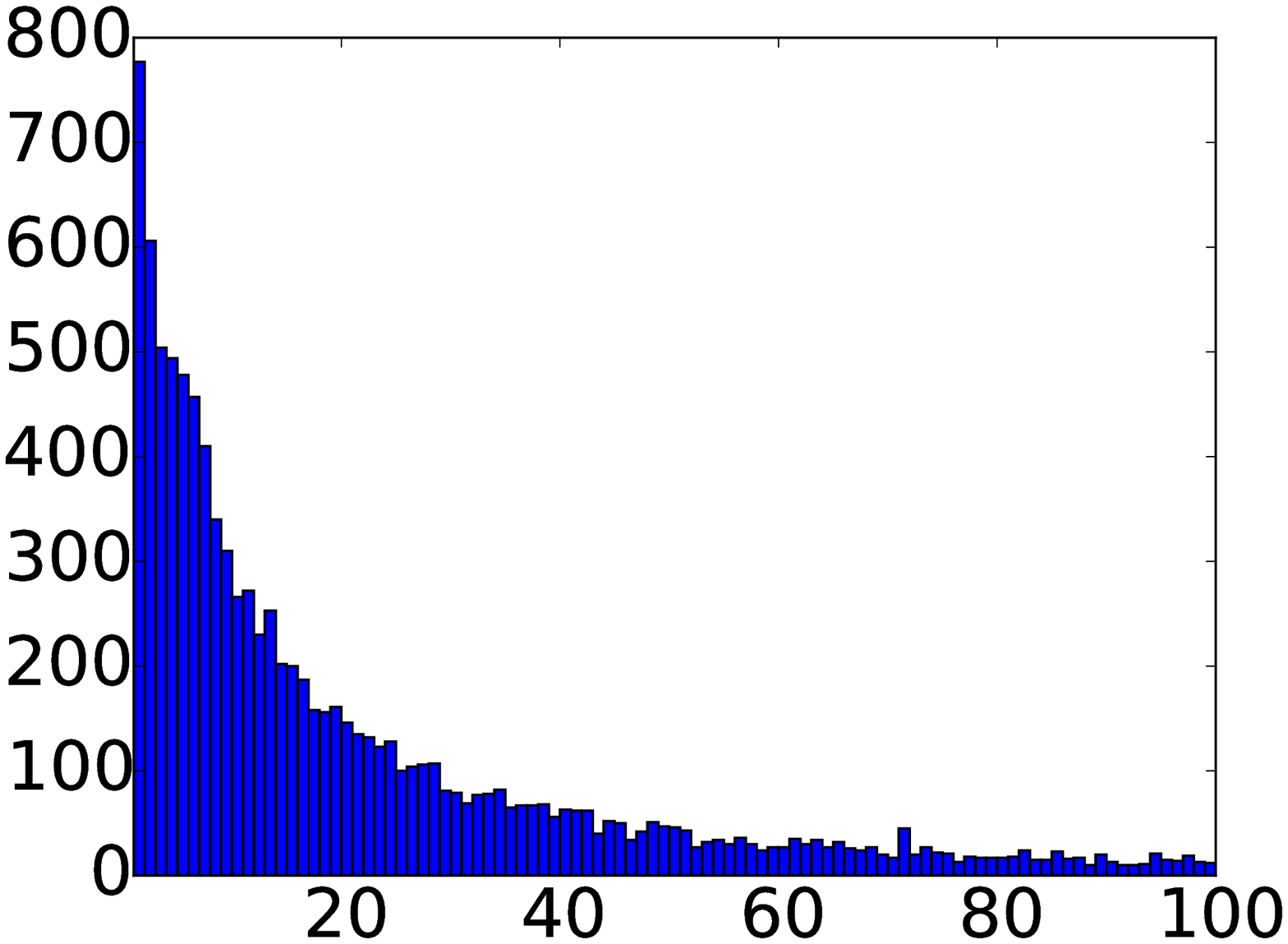}\tabularnewline
\hline 
\hline 
\multicolumn{3}{>{\centering}p{0.96\textwidth}}{\includegraphics[width=0.9\textwidth]{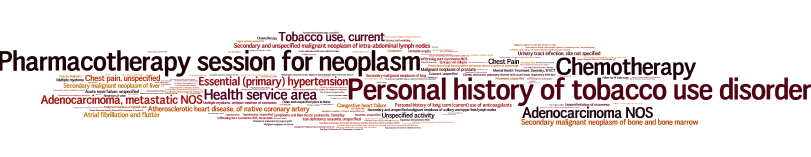} \includegraphics[width=0.9\textwidth]{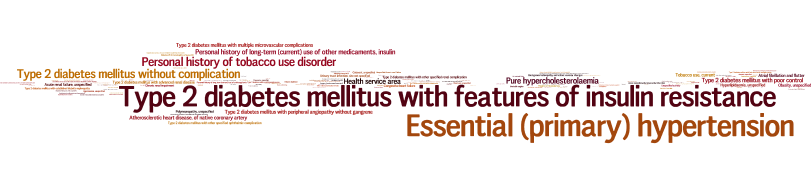}}\tabularnewline
\hline 
{\small{}{}Type 2 diabetes}{\small \par}

{\small{}{}Hypertension}  & {\small{}{}Tobacco use disorder}{\small \par}

{\small{}{}Type 2 w/o complication}  & {\small{}{}Long-term use of insulin}\tabularnewline
\hline 
\end{tabular}
\par\end{centering}

\protect\caption{\textbf{Top row}: Diabetes cohort statistics (y axis: number of patients;
x axis: (a) age, (b) number of admissions, (c) number of days);\textbf{
Mid row}: Progression from pre-diabetes (upper diag. cloud) to post-diabetes
(lower diag. cloud); \textbf{Bottom row}: Top diagnoses.\label{fig:diabetes_stats}}
\end{figure*}

\begin{figure*}
\begin{centering}
\begin{tabular}{>{\raggedright}p{0.3\textwidth}>{\raggedright}p{0.3\textwidth}>{\raggedright}p{0.3\textwidth}}
(a) Age  & (b) Admission  & (c) Length of stay (days)\tabularnewline
\includegraphics[width=0.3\textwidth]{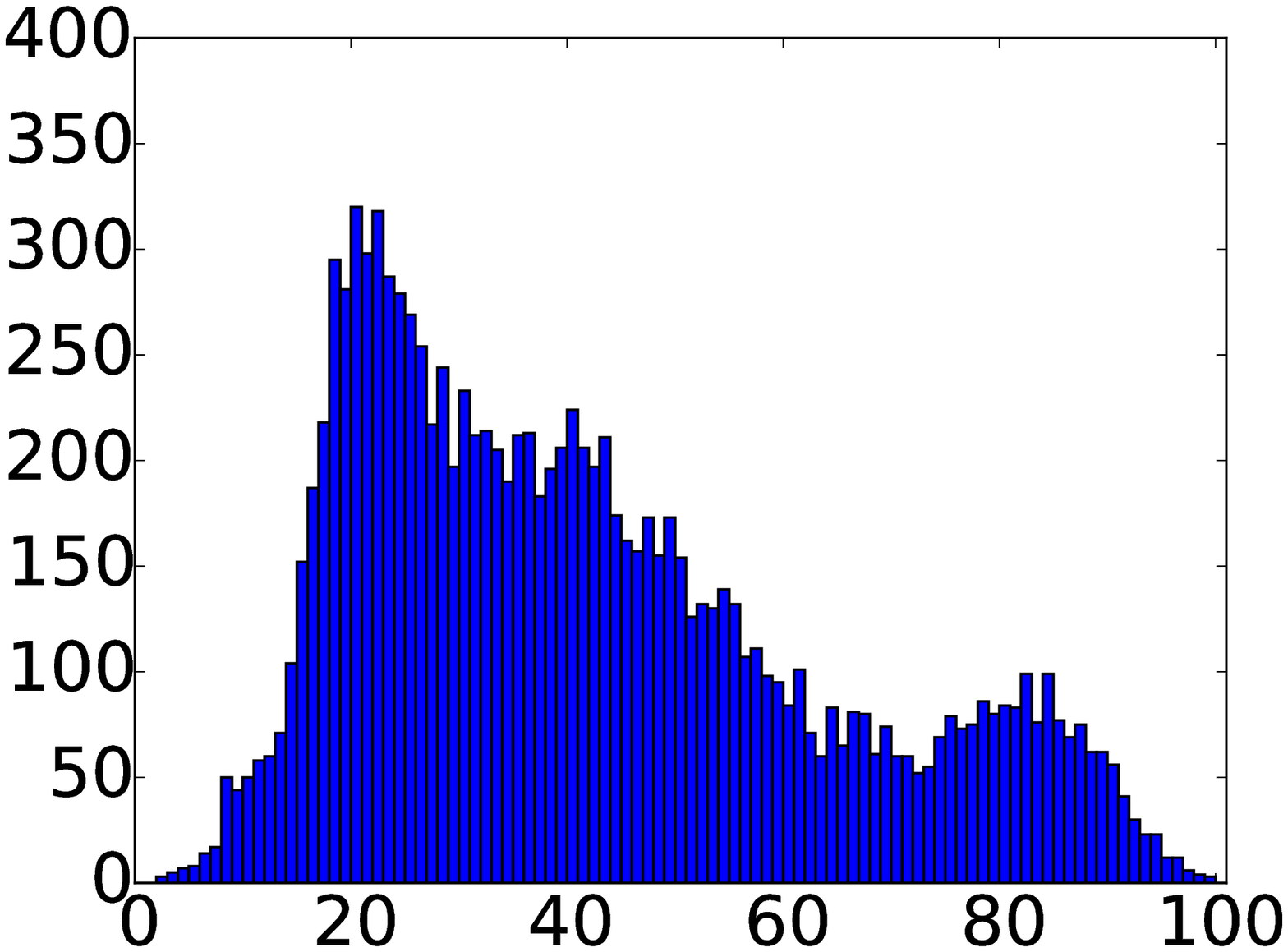}  & \includegraphics[width=0.3\textwidth]{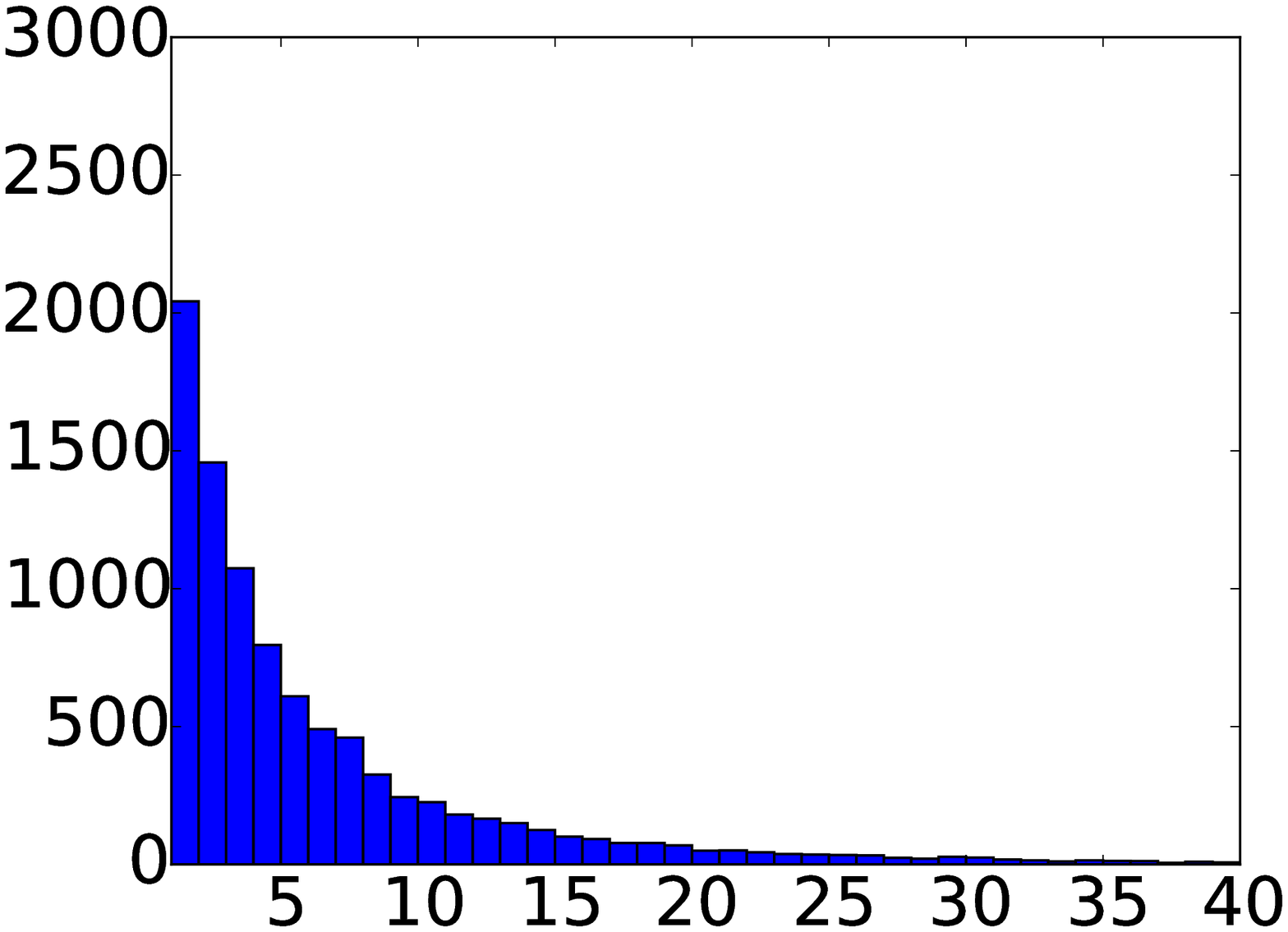}  & \includegraphics[width=0.3\textwidth]{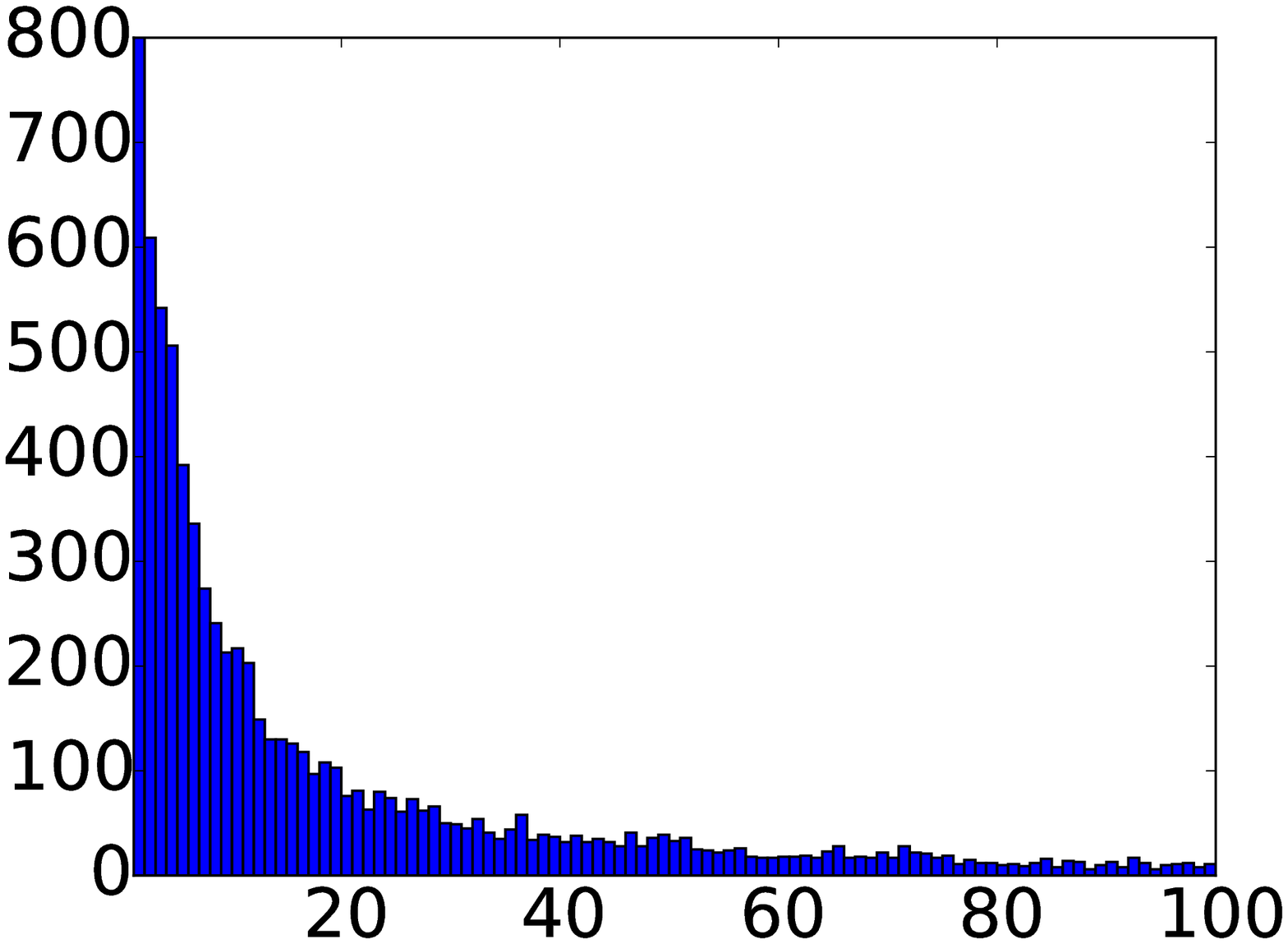}\tabularnewline
\hline 
\hline 
\multicolumn{3}{>{\centering}p{0.96\textwidth}}{\includegraphics[bb=0bp 195bp 812bp 340bp,clip,width=0.9\textwidth]{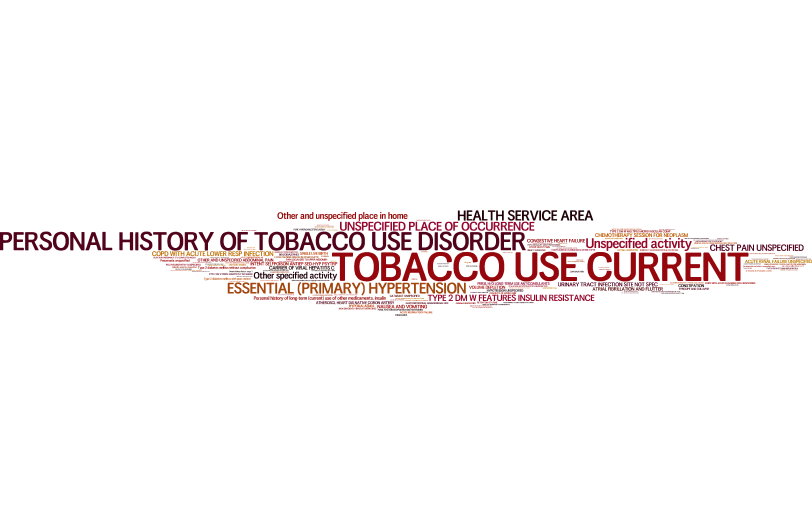}

\includegraphics[bb=0bp 195bp 812bp 340bp,clip,width=0.9\textwidth]{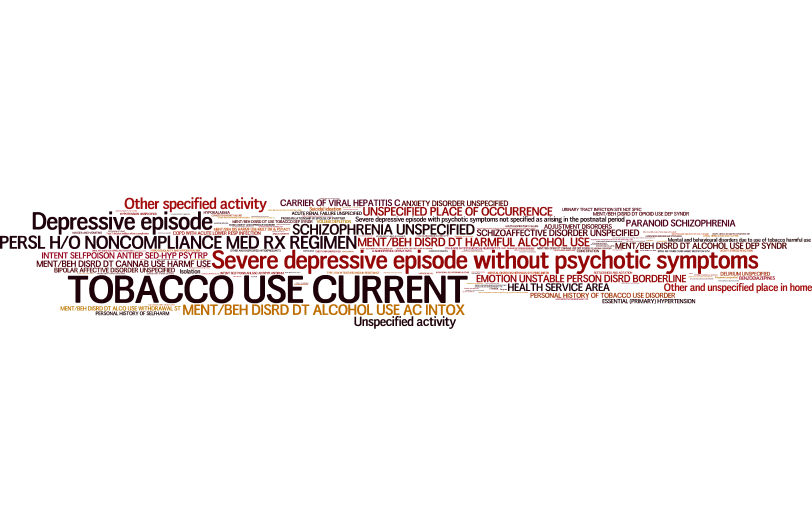}}\tabularnewline
\hline 
{\small{}{}Severe depressive episode}{\small \par}

Hypertension  & {\small{}{}Tobacco use current}{\small \par}

Mental/behavioral disorder due to alcohol use  & Personal history of tobacco use disorder\tabularnewline
\hline 
\end{tabular}
\par\end{centering}

\protect\caption{\textbf{Top row}: Mental health cohort statistics (y axis: number
of patients; x axis: (a) age, (b) number of admissions, (c) number
of days); \textbf{Mid row}: Progression from pre-mental diseases (upper
diag. cloud) to post-mental diseases (lower diag. cloud); \textbf{Bottom
row}: Top diagnoses.\label{fig:mental-stats}}
\end{figure*}

\subsection{Data}

Data for both cohorts were collected for 12 years (2002-2013) from
a large regional Australian hospital. We preprocessed the datasets
by removing (i) admissions with incomplete patient information; and
(ii) patients with less than 2 admissions. We define the vocabulary
as the set of diagnosis, procedure and medication codes. To reduce
the vocabulary, we collapse diagnoses that share the first 2 characters
into one diagnosis. Likewise, the first digits in the procedure block
are used.

The diabetes cohort contained more than 12,000 patients (55.5\% males,
median age 73). Data statistics are summarized in Fig.~\ref{fig:diabetes_stats}.
After preprocessing, the dataset contained 7,191 patients with 53,208
admissions. The vocabulary consisted of 243 diagnosis, 773 procedure
and 353 medication codes. The mental health cohort contains more than
11,000 patients (49.4\% males, median age 37). Data statistics are
summarized in Fig.~\ref{fig:mental-stats}. After preprocessing,
the mental health dataset contained 6,109 patients and 52,049 admissions
with the vocabulary of 247 diagnosis, 752 procedure and 319 medication
codes. The average age of diabetic patients is much higher than the
average age of mental patients (See Fig~\ref{fig:diabetes_stats}a
and Fig~\ref{fig:mental-stats}a).

\subsection{Implementation}

The training, validation and test sets are created by randomly dividing
the dataset into three parts of $2/3$, $1/6$, $1/6$ data points,
respectively. We vary the embedding and hidden dimensions from $5$
to $50$ but the results are rather robust. We report best results
for disease progression and intervention recommendation tasks with
$M=30$ and $K=40$ and for prediction tasks with $M=10$ embedding
dimensions and $K=20$ hidden units ($M$ and $K$ are the number
of embedding dimensions and hidden units respectively). Learning is
by Stochastic Gradient Descent with the mini-batch of $16$ sequences.
The learning rate $\lambda$ is modified as follows. We start with
$\lambda=0.01$. When the model cannot find a smaller training cost,
we wait $n_{wait}$ epochs before updating $\lambda$ as $\lambda=\lambda/2$.
Initially, $n_{wait}=5$, and is subsequently modified as $n_{wait}=\mbox{min}\left\{ 15,n_{wait}+2\right\} $
for each $\lambda$ update. Learning is terminated after $n_{epoch}=200$
or after learning rate smaller than $\epsilon=0.0001$.

\subsection{Disease progression \label{sub:Disease progression}}

We first verify that the recurrent memory embedded in DeepCare is
a realistic model of \emph{disease progression}. The model predicts
the next $n_{p}$ diagnoses at each discharge using Eq.~(\ref{eq:LSTM-label-pred}).

For comparison, we implement two baselines: Markov models and plain
RNNs. Markov model is a stochastic model used to model changing systems.
A Markov model consists of a list of possible states, the possible
transitions between those states and the probability of those transitions.
The future states depend only on the present state (Markov assumption).
The Markov model has memoryless disease transition probabilities $P\left(d_{t}^{i}\mid d_{t-1}^{j}\right)$
from disease $d^{j}$ to $d^{i}$ at time $t$. Given an admission
with disease subset $D_{t}$, the next disease probability is estimated
as $Q\left(d^{i};t\right)=\frac{1}{\left|D_{t}\right|}\sum_{j\in D_{t}}P\left(d_{t}^{i}\mid d_{t-1}^{j}\right)$.
Plain RNNs are described in Sec.~\ref{sub:Recurrent-Neural-Network}.

We use Precision at K (Precision@$K$) to measure the performance
of the models. Precision@$K$ corresponds to the percentage of relevant
results in retrieved results. That means if the model predicts $n_{p}$
diagnoses of the next readmission and $n_{r}$ diagnoses among of
them are relevant the model's performance is

\[
\mbox{Precision@}n_{p}=\frac{n_{r}}{n_{p}}
\]

\subsubsection*{Dynamics of forgetting}

\begin{figure}
\begin{centering}
\begin{tabular}{cc}
\includegraphics[bb=55bp 20bp 520bp 412bp,clip,width=0.4\columnwidth]{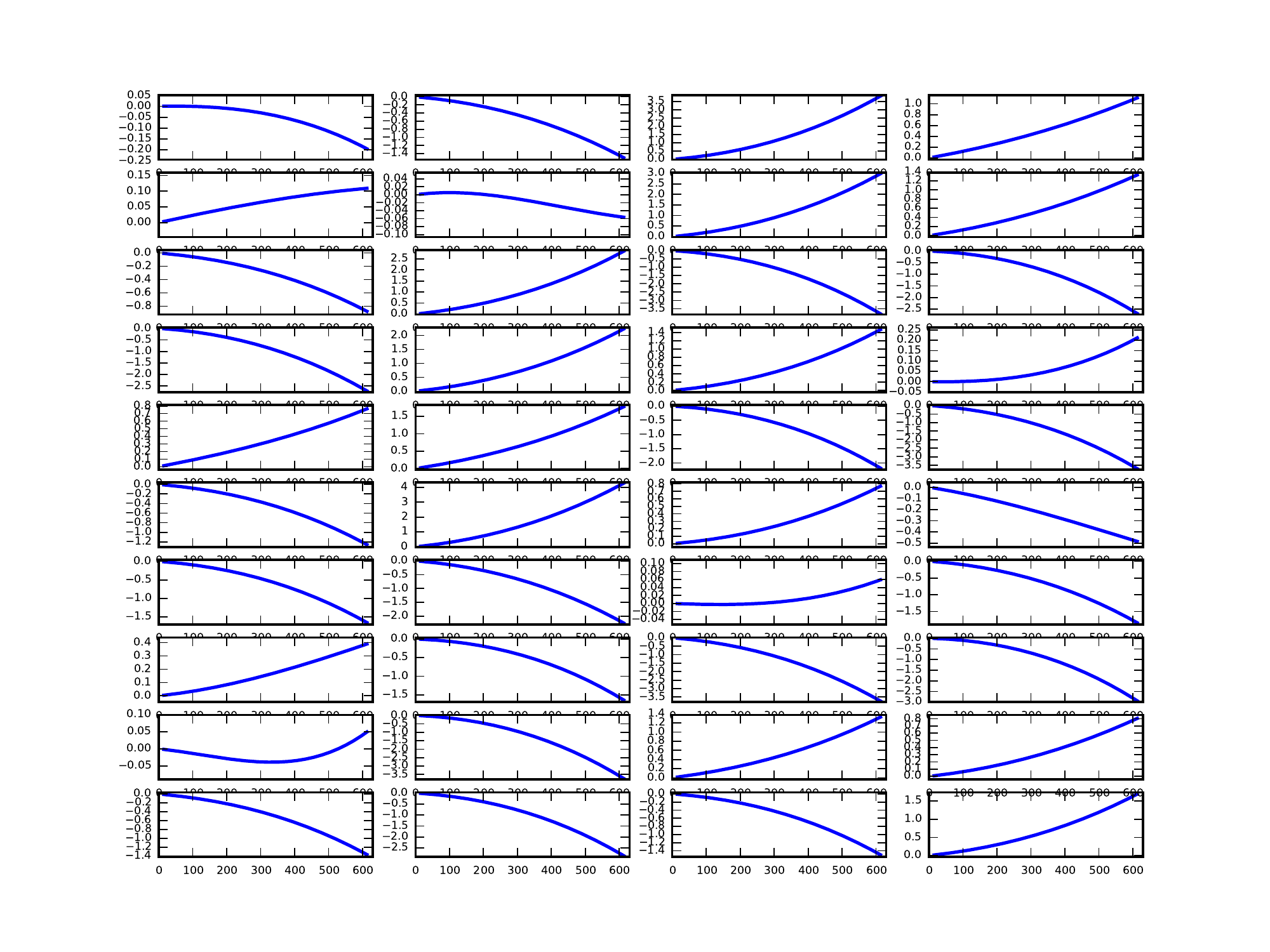}  & \includegraphics[bb=60bp 20bp 520bp 412bp,clip,width=0.4\columnwidth]{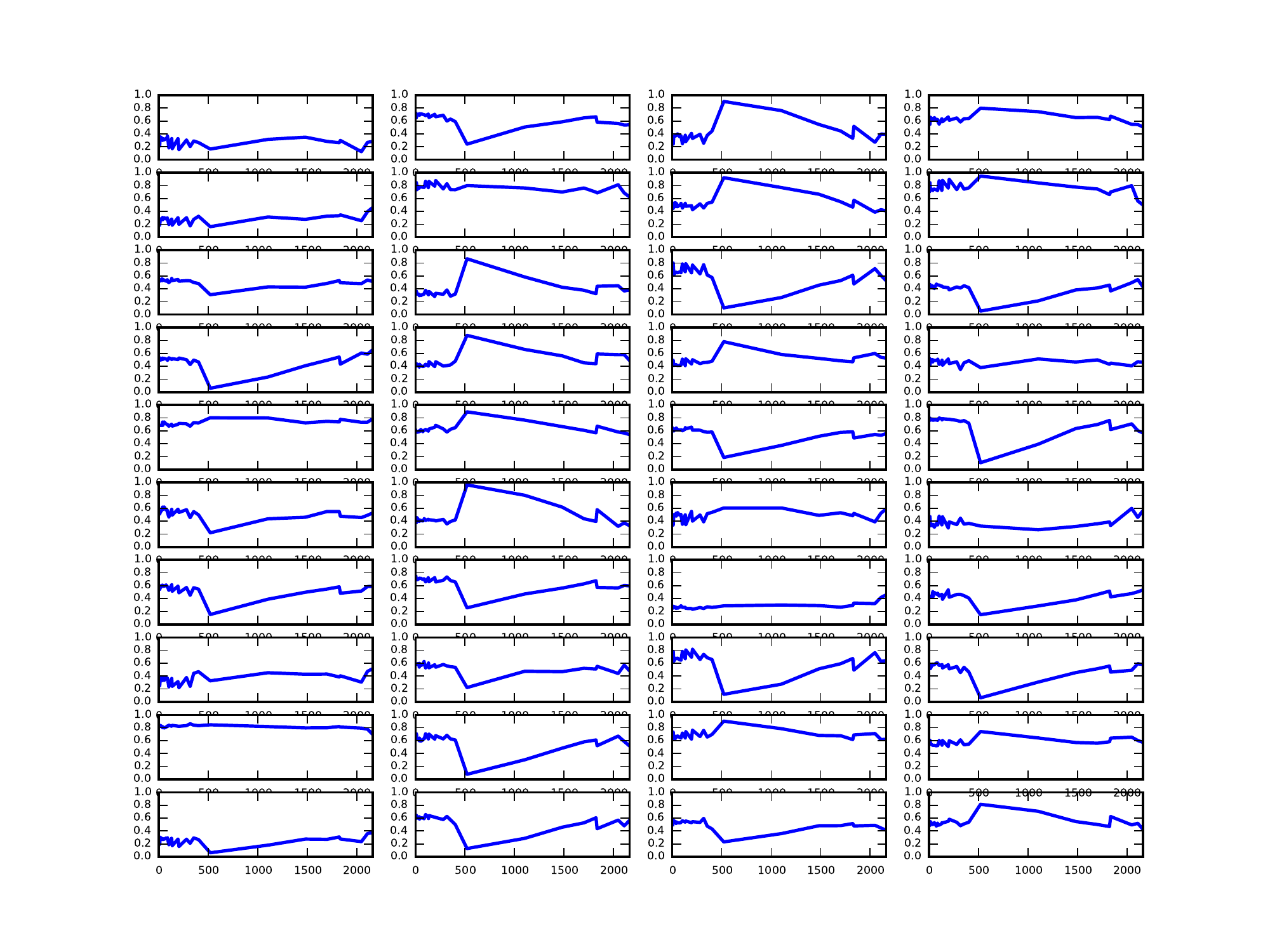}\tabularnewline
\end{tabular}
\par\end{centering}

\protect\caption{(Left) 40 channels of forgetting due to time elapsed. (Right) The
forget gates of a patient in the course of their illness. \label{fig:Forgetting-due-to-time}}
\end{figure}

Fig.~\ref{fig:Forgetting-due-to-time}(left) plots the contribution
of time into the forget gate. The contributions for all 40 states
are computed using $Q_{f}\qb_{\Delta_{t}}$ as in Eq.~(\ref{eq:time-parameterize}).
There are two distinct patterns: decay and growing. This suggests
that the time-based forgetting has a very small dimensionality, and
we will under-parameterize time using decay only as in Eq.~(\ref{eq:time-decay}),
and over-parameterize time using full parameterization as in Eq.~(\ref{eq:time-parameterize}).
A right balance is interesting to warrant a further investigation.
Fig.~\ref{fig:Forgetting-due-to-time}(right) shows the evolution
of the forget gates through the course of illness (2000 days) for
a patient.

\subsubsection*{Diagnoses prediction result}

Table~\ref{tab:disease-pred} reports the Precision@$n_{p}$ for
different values of $n_{p}$. For diabetes cohort, using plain RNN
improves over memoryless Markov model by $8.8\%$ with $n_{p}=1$
and by $27.7\%$ with $n_{pred}=3$. This significant improvement
demonstrates the role of modeling the dynamics in sequential data.
Modeling irregular timing and interventions in DeepCare gains a further
$2\%$ improvement. For mental health cohort, Markov model is failed
to predict next diagnoses with only $9.5\%$ for $n_{p}=1$. Plain
RNN gains $50\%$ improvement in Precision@1, while and DeepCare demonstrates
a $2\%$ improvement in Precision@1 over RNN.

\begin{table*}
\protect\caption{Precision@$n_{p}$ Diagnoses Prediction.\label{tab:disease-pred}}

\centering{}%
\begin{tabular}{l|ccc|ccc}
\cline{2-7} 
 & \multicolumn{3}{c|}{Diabetes} & \multicolumn{3}{c}{Mental}\tabularnewline
\cline{2-7} 
 & $n_{p}=1$  & $n_{p}=2$  & $n_{p}=3$  & $n_{p}=1$  & $n_{p}=2$  & $n_{p}=3$\tabularnewline
\hline 
\hline 
Markov  & 55.1  & 34.1  & 24.3  & 9.5  & 6.4  & 4.4\tabularnewline
Plain RNN  & 63.9  & 58.0  & 52.0  & 50.7  & 45.7  & 39.5\tabularnewline
\hline 
DeepCare (mean adm.)  & \textbf{66.2}  & \textbf{59.6}  & \textbf{53.7}  & \textbf{52.7}  & \textbf{46.9}  & \textbf{40.2}\tabularnewline
DeepCare (sum adm.)  & 65.5  & 59.3  & 53.5  & 51.7  & 46.2  & 39.8\tabularnewline
DeepCare (max adm.)  & 66.1  & 59.2  & 53.2  & 51.5  & 46.7  & \textbf{40.2}\tabularnewline
\hline 
\end{tabular}
\end{table*}

\subsection{Intervention recommendation \label{sub:Intervention-recommendation}}

Table~\ref{tab:Result-interventions} reports the results of current
intervention prediction. For all values of $n_{p}$, RNN consistently
outperforms Markov model by a huge margin for both diabetes and mental
health cohort. DeepCare with sum-pooling outperforms other models
in both diabetes and mental health datasets.

\begin{table*}
\protect\caption{Precision@$n_{p}$ intervention prediction\label{tab:Result-interventions}}

\centering{}%
\begin{tabular}{l|ccc|ccc}
\cline{2-7} 
 & \multicolumn{3}{c|}{Diabetes} & \multicolumn{3}{c}{Mental }\tabularnewline
\cline{2-7} 
 & $n_{p}=1$  & $n_{p}=2$  & $n_{p}=3$  & $n_{p}=1$  & $n_{p}=2$  & $n_{p}=3$\tabularnewline
\hline 
\hline 
Markov  & 35.0  & 17.6  & 11.7  & 20.7  & 12.2  & 8.1\tabularnewline
Plain RNN  & 77.7  & 54.8  & 43.1  & 70.4  & 55.4  & 43.7\tabularnewline
\hline 
DeepCare (mean adm.)  & 77.8  & 54.9  & 43.3  & 70.3  & 55.7  & 44.1\tabularnewline
DeepCare (sum adm.)  & \textbf{78.7}  & \textbf{55.5}  & \textbf{43.5}  & \textbf{71.0}  & \textbf{55.8}  & \textbf{44.7}\tabularnewline
DeepCare (max adm.)  & 78.4  & 55.1  & 43.4  & 70.0  & 55.2  & 43.9\tabularnewline
\hline 
\end{tabular}
\end{table*}

\subsection{Predicting future risk}

Next we demonstrate DeepCare on risk prediction. For each patient,
a discharge is randomly chosen as prediction point, from which \emph{unplanned
readmission} and \emph{high risk patients} within $X$ months will
be predicted. A patient is in high risk at a particular time $T$
if he or she have at least three unplanned readmissions within $X$
months after time $T$. We choose $X=12$ months for diabetes and
$X=3$ months for mental health.

For comparison,\textbf{ baselines} are SVM and Random Forests running
on standard non-temporal features engineering using one-hop representation
of diagnoses and intervention codes. Then pooling is applied to aggregate
over all existing admissions for each patient. Two pooling strategies
are tested: \emph{max} and \emph{sum}. Max-pooling is equivalent to
the presence-only strategy in \cite{arandjelovic2015discovering},
and sum-pooling is akin to an uniform convolutional kernel in \cite{tran2014framework}.
This feature engineering strategy is equivalent to zeros-forgetting
-- any risk factor occurring in the past is memorized.

\subsubsection*{Pretraining and Regularization}

Table~\ref{tab:Effect-of-pretraining} reports the impacts of pretraining
and regularization on the results of unplanned readmission prediction
in diabetes dataset using DeepCare model. Pretraining and regularization
improve the results of all three admission pooling methods. While
mean pooling admission is found to perform well with regularization,
max pooling produces best results with pretraining and sum pooling
produces best results with both approaches.

\begin{table}
\protect\caption{Effect of pretraining and regularization for unplanned readmission
prediction using DeepCare for diabetes dataset. The results are reported
in F-score (\%)\label{tab:Effect-of-pretraining}}

\centering{}%
\begin{tabular}{lccc}
\hline 
Approach  & Mean adm.  & Sum adm.  & Max adm.\tabularnewline
\hline 
\hline 
None  & 77.8  & 77.9  & 78.3\tabularnewline
\hline 
Pretrain  & 78.3  & 78.6  & \textbf{78.9}\tabularnewline
Regularization  & \textbf{79.0}  & 78.7  & 78.6\tabularnewline
\hline 
Both  & 78.4  & \textbf{78.9}  & 78.8\tabularnewline
\hline 
\end{tabular}
\end{table}

\subsubsection*{Unplanned readmission prediction results}

Table~\ref{tab:Prediction-results} reports the F-scores of predicting
unplanned readmission. For the diabetes cohort, the best baseline
(non-temporal) is Random Forests with \emph{sum pooling} has a F-score
of 71.4\% {[}Row 4{]}. Using plain RNN with simple logistic regression
improves over best non-temporal methods by a 3.7\% difference in 12-months
prediction {[}Row 5, ref: Sec.~(\ref{sub:Recurrent-Neural-Network},\ref{sub:Admission-Embedding}){]}.
Replacing RNN units by LSTM units gains 4.5\% improvement {[}Row 6,
ref: Sec.~\ref{sub:Long-Short-Term-Memory}{]}. Moving to deep models
by using a neural network as classifier helps with a gain of 5.1\%
improvement {[}Row 7, ref: Eq.~(\ref{eq:model}){]}. By carefully
modeling the irregular timing, interventions and recency+multiscale
pooling, we gain 5.7\% improvement {[}Row 8, ref: Secs.~(\ref{sub:Capturing-time-irregularity}--\ref{sub:Multiscale-pooling}){]}.
Finally, with parametric time we arrive at 79.0\% F-score, a 7.6\%
improvement over the best baselines {[}Row 9, ref: Secs.~(\ref{sub:Parametric-time}){]}.

For the mental health dataset, the best non-temporal baseline is \emph{sum-pooling
}Random Forest with result of 67.9\%. Plain RNN and LSTM with logistic
regression layer gain 2.6\% and 3.8\% improvements, respectively.
The best model is DeepCare with parametric time with a gap of 6.8\%
improvement compared to \emph{sum-pooling} Random Forest.

\begin{table*}
\protect\caption{Results of unplanned readmission prediction in F-score (\%) within
12 months for diabetes and 3 months for mental health patients (DC
is DeepCare, inv. is intervention).\label{tab:Prediction-results}}

\centering{}%
\begin{tabular}{lcc}
\hline 
Model  & Diabetes  & Mental\tabularnewline
\hline 
\hline 
1. SVM (\emph{max-pooling})  & 64.0  & 64.7\tabularnewline
2. SVM (\emph{sum-pooling})  & 66.7  & 65.9\tabularnewline
\hline 
3. Random Forests (\emph{max-pooling})  & 68.3  & 63.7\tabularnewline
4. Random Forests (\emph{sum-pooling})  & 71.4  & 67.9\tabularnewline
\hline 
5. Plain RNN (\emph{logist. regress.})  & 75.1  & 70.5\tabularnewline
6. LSTM (\emph{logit. regress.})  & 75.9  & 71.7\tabularnewline
\hline 
7. DC (\emph{nnets} + \emph{mean adm}.)  & 76.5  & 72.8\tabularnewline
8. DC ( {[}\emph{inv.+time decay}{]}\emph{+recent.multi.pool.+nnets}+\emph{mean
adm.})  & 77.1  & 74.5\tabularnewline
\textbf{9. DC ({[}}\textbf{\emph{inv.+param. time}}\textbf{{]}}\textbf{\emph{+recent.multi.pool.+nnets}}\textbf{+}\textbf{\emph{mean
adm.}}\textbf{)}  & \textbf{79.0}  & \textbf{74.7}\tabularnewline
\hline 
\end{tabular}
\end{table*}

\subsubsection*{High risk prediction results}

\begin{figure*}
\begin{centering}
\begin{tabular}{ccccc}
\includegraphics[bb=17bp 0bp 493bp 520bp,clip,width=0.35\textwidth]{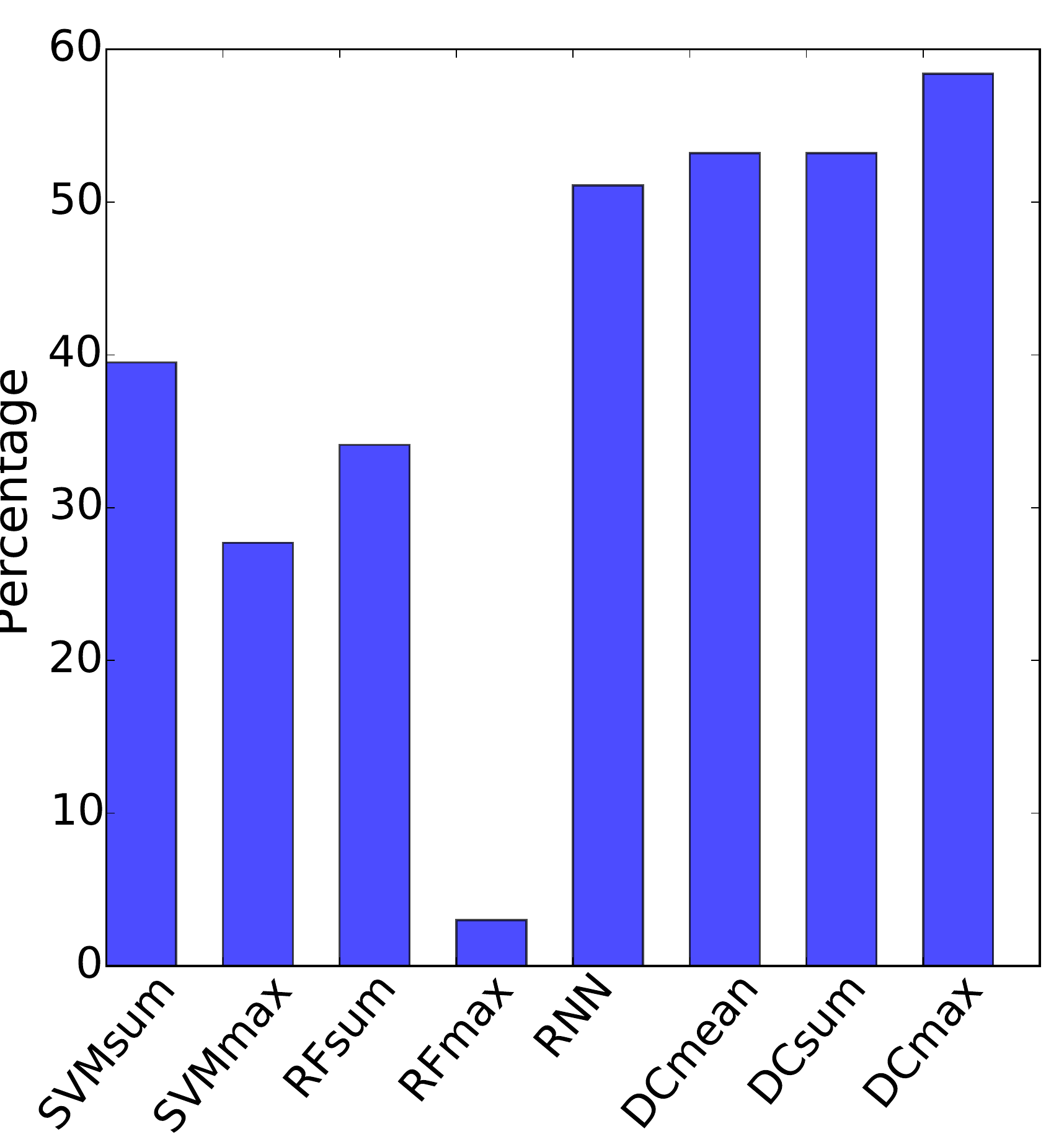}  &  &  &  & \includegraphics[bb=17bp 0bp 490bp 520bp,clip,width=0.35\textwidth]{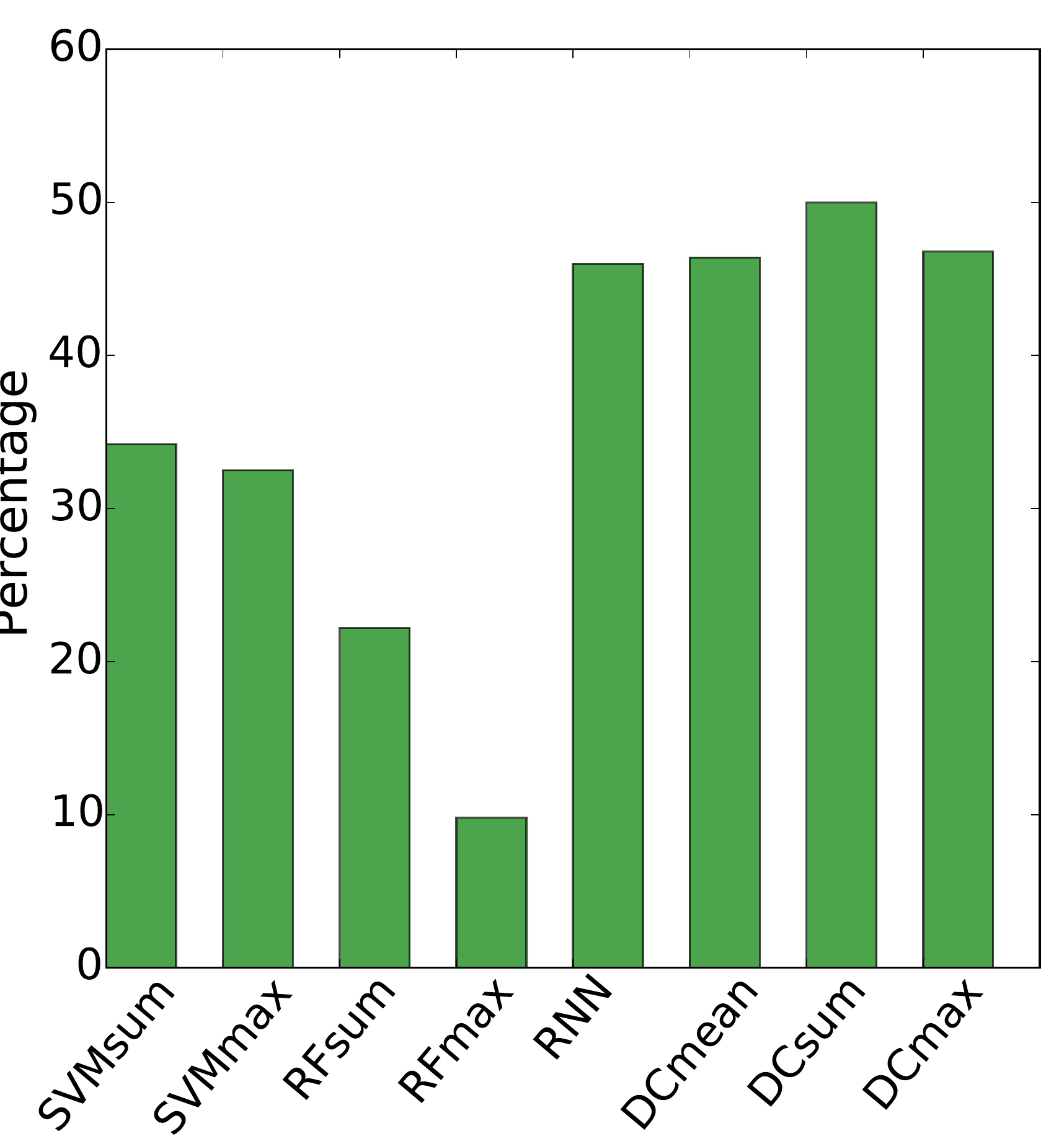}\tabularnewline
(a) Diabetes  &  &  &  & (b) Mental health\tabularnewline
\end{tabular}
\par\end{centering}

\protect\caption{Result of high risk prediction in F-score (\%) within 12 months for
diabetes (a) and 3 months for mental health (b). DC is DeepCare. Mean,
sum, max are 3 admission pooling methods \label{fig:Result-of-high-risk}}
\end{figure*}

In this part, we report the performance of DeepCare on high risk patient
prediction task. Figure~\ref{fig:Result-of-high-risk} reports the
F-score of high risk prediction. RNN improves the best non-temporal
model (\emph{sum-pooling} SVM) over 10\% F-score for both two cohorts.
Max-pooling DeepCare best performs in diabetes dataset with nearly
60\% F-score, while sum-pooling DeepCare wins in mental health cohort
with 50.0\% F-score.

\section{Discussion and Conclusion \label{sec:Discussion-and-Conclusion}}

\subsection{Discussion}

DeepCare was partly inspired by human memory \cite{baddeley1992working}.
There are three kinds of related memory: semantic, episodic and working
memory. \emph{Semantic} \emph{memory} stores the meaning of concepts
and their relations. \emph{Episodic} \emph{memory} refers to the storage
of experiences triggered by an event, for example, wedding or earthquake.
\emph{Working} memory is a system of temporarily loading and processing
information as part of complex cognitive tasks.

DeepCare makes use of embedding to represent the semantics of diagnoses,
interventions and admissions. In theory, this embedding can be estimated
independently of the task at hand. Our previous work learns diagnosis
and patient embedding \cite{tran2015learning} using nonnegative restricted
Boltzmann machines \cite{tu_truyen_phung_venkatesh_acml13} and known
semantic relations and temporal relations \cite{nguyen2016graph}.
This method uses global contexts, unlike DeepCare, where only local
contexts (e.g., next admission) are considered.

The memory cells in DeepCare are used to store, update, forget and
manipulate illness experiences over time-stamped episodes. The inferred
experiences are then pooled to reason about the current illness states
and the future prognosis. Like human memory, healthcare risk also
has a recency effect, that is, more recent events contribute more
into the future risk. In DeepCare, two recency mechanisms are used.
First, through forgetting, recent events in DeepCare tend to contribute
more to the current illness states. Second, multiscale pooling as
in Sec.~\ref{sub:Multiscale-pooling} has weights decayed over time.

DeepCare can be implemented on existing EMR systems. For that more
extensive evaluations on a variety of cohorts, sites and outcomes
will be necessary. This offers opportunities for domain adaptations
through parameter sharing among multiple cohorts and hospitals. Modeling-wise,
DeepCare can also be extended to predict a sequence of outcomes at
specific timing, in the same spirit as the sequence to sequence mapping
in \cite{sutskever2014sequence}. Future work also includes more flexibility
in time parameterization such as using radial basis expansion and
splines. Further, DeepCare is generic so it can be applied to not
only medical data but also other kinds of sequential data which contain
long-term dependencies, sequence of sets, irregular time and interventions.

\subsection{Conclusion}

In this paper we have introduced DeepCare, an \emph{end-to-end} deep
dynamic memory neural network for personalized healthcare. It frees
model designers from manual feature extraction. DeepCare reads medical
records, memorizes illness trajectories and care processes, estimates
the present illness states, and predicts the future risk. Our framework
models disease progression, supports intervention recommendation,
and provides prognosis from electronic medical records. To achieve
precision and predictive power, DeepCare extends the classic Long
Short-Term Memory by (i) embedding variable-size discrete admissions
into vector space, (ii) parameterizing time to enable irregular timing,
(iii) incorporating interventions to reflect their targeted influence
in the course of illness and disease progression; (iv) using multiscale
pooling over time; and finally (v) augmenting a neural network to
infer about future outcomes. We have demonstrated DeepCare on predicting
next disease stages, recommending interventions, and estimating unplanned
readmission among diabetic and mental health patients. The results
are competitive against current state-of-the-arts. DeepCare opens
up a new principled approach to predictive medicine.

\bibliographystyle{IEEEtran}
\bibliography{ME,truyen}

\end{document}